\documentclass[journal]{IEEEtai}
\usepackage[colorlinks,urlcolor=blue,linkcolor=blue,citecolor=blue]{hyperref}
\usepackage{color,array}
\IEEEoverridecommandlockouts
\usepackage{cite}
\usepackage{amsmath,amssymb,amsfonts}
\usepackage{algorithmic}
\usepackage{textcomp}
\def\BibTeX{{\rm B\kern-.05em{\sc i\kern-.025em b}\kern-.08em
		T\kern-.1667em\lower.7ex\hbox{E}\kern-.125emX}}

\usepackage{subcaption}
\usepackage{algorithm}
\usepackage{url}
\usepackage{stfloats}

\usepackage{textcomp}
\usepackage{siunitx}
\usepackage{float}

\usepackage{rotating}
\usepackage{multirow}
\usepackage{booktabs}
\usepackage{bigstrut}

\usepackage{clipboard}
\usepackage{xcolor, soul} 
	
\definecolor{OliveGreen}{cmyk}{0.64,0,0.95,0.40}
\definecolor{marygold}{cmyk}{0,0.1,0.5,0}

\usepackage{balance}
\usepackage{multicol}
\usepackage{flushend}

\usepackage{mathtools}
\usepackage{cuted}

\newcommand{\MethodnameLong}{Synthetic Information towards Maximum Posterior Ratio}
\newcommand{\Methodname}{SIMPOR}

\begin{document}

\title{\MethodnameLong{} for deep learning on Imbalanced Data\\}

\author{\IEEEauthorblockN{Hung Nguyen\IEEEauthorrefmark{1}
		and~J. Morris Chang\IEEEauthorrefmark{2}}
	\IEEEauthorblockA{Department of Electrical Engineering \\
		University of South Florida \\ Tampa, Florida 33620\\
		Email: \IEEEauthorrefmark{1}nsh@usf.edu, \IEEEauthorrefmark{2}chang5@usf.edu}}

\maketitle
\thispagestyle{plain}
\pagestyle{plain}

\begin{abstract}

	This work explores how class-imbalanced data affects deep learning and proposes a data balancing technique for mitigation by generating more synthetic data for the minority class. In contrast to random-based oversampling techniques, our approach prioritizes balancing the most informative region by finding high entropy samples. This approach is opportunistic and challenging because well-placed synthetic data points can boost machine learning algorithms' accuracy and efficiency, whereas poorly-placed ones can cause a higher misclassification rate. In this study, we present an algorithm for maximizing the probability of generating a synthetic sample in the correct region of its class by placing it toward maximizing the class posterior ratio. In addition, to preserve data topology, synthetic data are closely generated within each minority sample neighborhood. Overall, experimental results on forty-one datasets show that our technique significantly outperforms experimental methods in terms of boosting deep-learning performance. 
\end{abstract}

\begin{IEEEImpStatement}
	Data class imbalance is a well-known problem in machine learning (ML) applications. This significantly reduces ML algorithms' performance because models are biased toward the majority class. While several strategies have been proposed to mitigate the problem for traditional ML, there is a lack of research for deep learning. In contrast to rule-based ML algorithms, deep learning is highly data-dependent, so understanding how a deep model is affected by data is crucial for finding the mitigations. We provide intuitive studies of different mitigation strategies on deep learning models to fill this gap. Besides a minority oversampling-based technique is proposed to address the problem, a combination of a heuristic technique to find high entropy samples and a conventional statistical theorem to determine where synthetic samples should be spawned. Because our technique is directly designed to tackle the issue of class imbalance for deep learning models, it has been shown to achieve the highest number of winning times (in two metrics, F1-score and AUC) over 41 real datasets compared to the other techniques. The Wilcoxon signed-rank test also shows the significance of the improvement.   
\end{IEEEImpStatement}


\section{Introduction}
Class imbalance is a common phenomenon; it could be caused by the data collecting procedure or simply the nature of the data. For example, it is difficult to sample some rare diseases in the medical field, so collected data for these are usually significantly less than that for other diseases. This leads to the problem of class imbalance in machine learning. The chance of rare samples appearing in model training process is much smaller than that of common samples. Thus, machine learning models tend to be dominated by the majority class; this results in a higher prediction error rate. Existing work also observed that class imbalanced data cause a slow convergence in the training process because of the domination of gradient vectors coming from the majority class \cite{ya-guan_emsgd:_2020, liu_high-performance_2020}. 

In the last decades, a number of techniques have been proposed to soften the negative effects of class imbalance for conventional machine learning algorithms by analytically studying particular algorithms and developing corresponding strategies. However, the problem for heuristic algorithms such as deep learning is often more difficult to tackle. As suggested in the most recent deep learning with class imbalance survey \cite{johnson_survey_2019}, most of the works are emphasizing image data, and studies for other data types are missing. Thus, in this work, we focus on addressing the issue of tabular data with class imbalance for deep learning models. A class balancing solution is proposed that utilizes entropy-based sampling and data statistical information. As suggested in the survey (\cite{johnson_survey_2019}) that techniques for traditional ML can be extended to deep learning and inspiring by the comparison in a recent relevant work, Gaussian Distribution Based Oversampling (GDO) \cite{bib:GDO}, we compare the proposed technique with other widely-used and recent techniques such as GDO \cite{bib:GDO}, SMOTE \cite{chawla_smote:_2002}, ADASYN \cite{ADASYN}, Borderline SMOTE \cite{bordersmote}, DeepSMOTE \cite{deepsmote}. 

Current solutions can be classified into two approaches: model-centric and data-centric. The former strives to alter machine algorithms, while the latter focuses on finding data balancing methods. Perhaps data-centric techniques are more commonly used because they do not tie to any specific model. In this category, a simple data balancing technique is to duplicate minority instances to balance the sample quantity between classes, namely random oversampling (ROS). This can preserve the best data structure and reduce the negative impact of data imbalance to some degree. However, this puts too much weight on a very few minority samples; as a result, it causes over-fitting problems in deep learning when the imbalance ratio becomes higher.

Another widely-used technique in this category is Synthetic Minority Oversampling Technique (SMOTE) \cite{chawla_smote:_2002}, which randomly generates synthetic data on the connections (in Euclidean space) between minority samples. However, this easily breaks data topology, especially in high-dimensional space, because it can accidentally connect instances that are not supposed to be connected. In addition, if there are minority samples located in the majority class, the technique will generate sample lines across the decision boundary, which leads to distorted decision boundaries and misclassification. To improve SMOTE, Hui Han, \textit{et al.} \cite{bordersmote} proposed a SMOTE-based technique (Borderline SMOTE), in which they only apply SMOTE on the near-boundary samples determined by the labels of their neighbors. Since this technique is entirely based on Euclidean distance from determining neighbors to generating synthetic data, it performs poorly in high dimensional space. To enhance oversampling with high dimensional data such as images, Dablain \textit{et al.} introduced DeepSMOTE \cite{deepsmote} in 2022 which is a combination of SMOTE and a GAN (generative adversarial network). Similar to SMOTE, if there is any poorly generated sample near the boundary, it will worsen the problem due to synthetic samples bridges across the border. Leveraging the same way as SMOTE generates synthetic samples, another widely-used technique, ADASYN \cite{ADASYN}, controls the number of generated samples by the number of samples in different classes within small groups. Again, this technique still suffers distortion of the decision boundary if the boundary region is class imbalanced. Additionally, such mentioned techniques have not utilized statistical data information. A recent work, Gaussian Distribution Based Oversampling (GDO) \cite{bib:GDO}, balances data class based on the statistical information of data instead. However, its strong assumption of data distribution (data follow Gaussian) reduces the technique's effectiveness in real data.  

To alleviate the negative effects of data imbalance and avoid the drawbacks of existing techniques, a minority oversampling technique is proposed that focuses on balancing the high-entropy region that provides the most critical information to the deep learning models. Besides, the technique enhances synthetic data's chance to fall into the minority class to reduce model errors. By carefully generating synthetic data near minority samples, our proposed technique also preserves the best data topology. Besides, our technique does not need any statistical assumption. 

To find informative samples, an entropy-based deep active learning technique is used to select samples yielding high entropy to deep learning models. The region of informative samples is denoted as the informative region. This region is balanced first, and the remaining data are balanced later so that it would reduce the decision distortion mentioned earlier. For each minority sample in this region, its synthetic neighbors are safely generated so that the global data topology is still preserved. However, generating synthetic samples in this region is risky because it can easily fall across the decision boundary. Therefore, a direction for synthetic sample location can be chosen by maximizing its posterior probability based on Bayes's Theorem. However, maximizing the posterior probability is facing infeasible computation in the denominator. To overcome this, the posterior ratio is maximized instead so that the denominator computation can be avoided. This also ensures that the synthetic samples are not only close to the minority class but also far from the majority class. The remaining data are eventually balanced by a similar procedure. 

The proposed technique alleviates the class imbalance problem. Overall, our experiments indicate that the proposed method can achieve better classification results over widely-used techniques.

Our work has the following main contributions:
\begin{enumerate}
	\item{Exploring the impact of class imbalance mitigations on deep learning via visualization and experiments.}
	\item{Proposing a new minority oversampling-based technique, namely \MethodnameLong, to balance data classes and alleviate data imbalance impacts. Our technique is enhanced by the following key points.}
	\begin{enumerate}
		\item Leveraging an entropy-based active learning technique to prioritize the region that needs to be balanced. It is the informative region where samples provide high information entropy to the model. 
		\item Leveraging Maximum Posterior Ratio and Bayes's theorem to determine the direction to generate synthetic minority samples to ensure the synthetic data fall into the minority class and not fall across the decision boundary. To our best knowledge, this is the first work utilizing the posterior ratio for tackling class imbalanced data. 
		\item Approximating the likelihood in the posterior ratio using kernel density estimation, which can approximate a complicated topology. Thus, the proposed technique is able to work with large, distributively complex data. 
		\item Carefully generating synthetic samples surrounding minority samples so that the global data topology is still preserved. 
	\end{enumerate}
	\item{The proposed technique is evaluated against commonly utilized and contemporary techniques across 41 actual datasets that vary in terms of imbalance ratio and feature count. The findings demonstrate that the proposed approach exhibits superior performance compared to others, on the whole.}
\end{enumerate}

The rest of this paper is organized as follows. Section \ref{sec:relatedwork} briefly review other existing works. Section \ref{sec:preliminaries} introduces related concepts that will be used in this work, i.e., Imbalance Ratio, Macro F1-score, Area Under the Curve (AUC), and Entropy-based active learning. Section \ref{sec:problem} will provide more detail on the problem of learning from an imbalanced dataset. Our proposed solution to balance dataset, \MethodnameLong, will be explained comprehensively in Section \ref{sec:SIMPOR_method}. Section \ref{sec:implementation} discusses the technique implementation and complexity.  Section \ref{sec:experiments} shows experiments on different datasets, including artificial and real datasets. Experimental results are also discussed in the same section. Section \ref{sec:conclusion} concludes the study and discusses future work.

\section{Related Work}
\label{sec:relatedwork}
In the last few decades, many solutions have been proposed to alleviate the negative impacts of data imbalance in machine learning. However, most of them are not efficiently extended for deep learning. This section reviews algorithms to tackle class-imbalanced data that can be extended for deep learning. These techniques can be categorized into three main categories, i.e., sampling, cost-sensitive, and ensemble learning approaches. 

\subsection{Sampling-based approach.}
Compared to other approaches, resampling techniques have attracted more research attention as they are independent of machine learning algorithms. This approach can be divided into two main categories, over-sampling, and under-sampling techniques. Such sampling-based methods e.g., \cite{DBLP:journals/corr/ShenLH15, DBLP:journals/corr/abs-1711-00941, haibo_he_learning_2009, li_entropy-based_2020,ertekin_active_2007,8745666,8713384} mainly generate a balanced dataset by either over-sampling the minority class or down-sampling the majority class. Liu \textit{et al.}, \cite{9723474, 8668459} proposed two different approaches to learn from imbalanced data by capturing critical features in minority examples using model-based and density based methods. Some techniques are not designed for deep learning; however, they are still considered in this work since they are independent of the machine learning model architecture. In a widely used method, SMOTE \cite{chawla_smote:_2002}, Chawla \textit{et al.} attempt to oversample minority class samples by connecting a sample to its neighbors in feature space and arbitrarily drawing synthetic samples along with the connections. However, one of SMOTE drawbacks is that if there are samples in the minority class located in the majority class, it creates synthetic sample bridges toward the majority class \cite{goswami_class_2020}. This renders difficulties in differentiation between the two classes. Another SMOTE-based work, namely Borderline-SMOTE \cite{bordersmote} was proposed in which its method aims to do SMOTE with only samples near the border between classes. The samples near the border are determined by the labels of its \textit{k} distance-based neighbors. This "border" idea is similar to ours to some degree. However, finding a good \textit{k} is critical for a heuristic machine learning algorithm such as deep learning, and it is usually highly data-dependent. 

Among specific techniques for deep learning, generating synthetic samples in the minority class by sampling from data distribution is becoming more attractive as they outperform other methods in high dimensional data \cite{DBLP:conf/dmin/LiuGM07}. Regarding images, several deep learning generative-based methods have been proposed as deep learning is capable of capturing good image representations. \cite{rashid_convergence_2012} \cite{dai_generative_2019} \cite{mullick_generative_2019} utilized Variational Autoencoder as a generative model to arbitrarily generate images from learned distributions. However, most assumed simple prior distributions, such as Gaussian for minor classes, tend to simplify data distribution and might fail in more sophisticated distributions. In addition, most of the works in this approach tackle image datasets, while our proposed method focuses on tabular datasets as this is a missing piece in the field \cite{johnson_survey_2019}.

Under the down-sampling category, existing techniques mainly down-sample the majority class to balance it with the minority class. There are several proposed techniques to simplify the majority. A straightforward way is to randomly remove the majority class samples. Other works, e.g., \cite{ertekin_learning_2007}, \cite{aggarwal_active_2020} find near-border samples and authors believe the imbalance ratio in these areas is much smaller than that in the entire dataset. They then classify this small pool of samples to improve the performance and expedite the training process for the SVM-based method. However, this method was only designed for SVM-based methods, which mainly depend on the support vectors. Also, this potentially discards essential information of the entire dataset because only a small pool of data is used.  

\subsection{Cost-sensitive learning approach.}
Cost-sensitive learning techniques usually require modifications of algorithms on the cost functions to balance each class's weight. Specifically, such cost-sensitive techniques put higher penalties on majority classes and less on minority classes to balance their contribution to the final cost. For example,  \cite{cui_class-balanced_2019} provided their designed formula $ (1 - \beta^n)/(1 - \beta)$ to compute the weight of each class based on the effective number of samples $n$ and a hyperparameter $\beta$ which is then applied to re-balance the loss of a convolutional neural network model.  \cite{huang_learning_2016},\cite{rangarajan_sridhar_unsupervised_2015}, \cite{DBLP:journals/corr/abs-1805-00932} assign classes' weights inversely proportional to sample frequency appearing in each class. Hamed et al. \cite{cssvm} proposed an SVM-based cost-sensitive approach (SVMCS) that uses svm with a class-weighted loss function.  
  
\subsection{Ensemple learning approach.}
Ensemble learning has achieved high performance in classification for its generalizability. Thus, it could reduce the bias due to class imbalance. Ensemble learning can be constructed by combining several base classifiers with different sampling-based approaches. In \cite{SMOTEBoost, RUSBoost}, Chawla \textit{et al.} and Seiffert \textit{et al.} proposed variants of ensemble learning in which the data are balanced based on oversampling method SMOTE and then applying ensemble learning on balanced data. Similarly, authors in \cite{ECOEnsemble} generate cluster-based synthetic data and combine it with an evolutionary algorithm. Liu \textit{et al.} in \cite{LIU201735} balances the data by applying a fuzzy-based oversampling technique and building ensemble learning classifiers on this data. Zhou and Liu in \cite{EE} explore a method, namely Easy Ensemble classifier (EE), to perform ensemble learning on the random under-sampling balanced data.

\section{Preliminaries}
\label{sec:preliminaries}       
In this section, we introduce relevant concepts that will be utilized in our research.

\subsection{Imbalance Ratio (IR)}
For binary classification problems, imbalance ratio (IR) is used to depict the data imbalance as it has been widely used. IR is the ratio of the majority class samples to the minority class's samples. For example, if a dataset contains 1000 class-A and 100 class-B samples, the Imbalance Ratio is 10:1.   

\subsection{Evaluation Metrics}
\label{f1score}
In this work, classification performance is used for evaluating techniques. Specifically, F1-Score and Area Under the Curve (AUC) are used for evaluation metrics. For F1-scores, Macro-averaging is measured as it is more relevant for evaluating imbalance datasets.
F1 score is computed based on two factors Recall and Precision as follows:\\
\begin{align}
	Recall &= \frac{TP}{TP+FN}\\
	Precision &= \frac{TP}{TP+FP}\\
	F1-score &= \frac{2*Recall*Precision}{Recall+Precision},
\end{align}
where $T$ and $F$ stand for True and False; $P$ and $N$ stand for Positive and Negative. 

Besides, AUC \cite{cite:AUC} score is computed as it is an important metric to evaluate imbalanced data. AUC is derived from the Receiver Operating Characteristic curve (ROC). In this work, a skit-learn library to compute AUC is utilized; the library can be found in sklearn.metrics.auc. 

\subsection{Entropy-based Active Learning }   
\label{sec:EAL}
Entropy-based active learning (AL) \cite{shannon_mathematical_1948} is leveraged to find informative samples. The technique selects samples that provide high information to the model based on information entropy theory. The information entropy is quantified based on the ``surprise" to the model in terms of class prediction probability. Take a binary classification, for example; if a sample is predicted to be 50\% belonging to class A and 50\% belonging to class B, this sample has high entropy and is informative to the model. In contrast, if it is predicted to be 100\% belonging to class A, it is certain and gives zero information to the model. The class entropy $E$ for each sample can be computed as follows. 

\begin{equation}
	E(x,\theta) = -\sum_j^n{ P_\theta( y=c_j|x) \log_n P_\theta(y=c_j|x) }
	\label{eq:entropy_AL}
\end{equation} 
where $P_\theta(y=c_j|x)$ is the probability of data $x$ belonging to the $j^{th}$ class of $n$ classes with the model parameter $\theta$.

\Copy{EntropyActiveLearning}{
To select informative samples, one can fully train the entire original dataset and estimate entropy scores on the same data to select high-entropy samples. However, the model fully training on the entire dataset might be biased due to the data imbalance. This could lead to a bias in selecting informative samples because the model might only recognize the densest minority area and ignore other areas containing fewer informative examples. To avoid this issue, we proposed to explore more informative samples batch by batch gradually; this mechanism was inspired by the idea of exploring critical data by batches from active learning. First, the model is trained with an initial batch of data. The model is then used to estimate entropy scores for the remaining unseen data to select the first set of high-entropy samples (this set is considered informative examples relative to the current model parameters). This high informative data is then accumulated to previous training data to continue fine-tuning the current model and select the next informative set from the rest of the data. The process is repeated until reaching the desired amount of informative samples. The remainder of this section describes more detail on the mechanism.

The proposed approach implementation requires repeated phases, and a batch of informative data is selected for each phase. At the first phase $t^{(0)}$, a classifier with parameter $\theta^{(0)}$ (Note that this classifier differs from the classifier for the final classification problem) is trained on an initial batch of data (at least one sample in each class is required) and use the model $\theta^{(0)}$ to estimate the entropy for the remaining data. The entropy scores are then estimated for the remaining samples based on Equation \ref{eq:entropy_AL}. The first batch of informative samples is determined by selecting $k$ highest entropy samples. This batch is then concatenated with the initial training data for the training classifier parameter ($\theta^{(1)}$) in the next phase ($t^{(1)}$) and also accumulated to the informative set. In the next phase, similarly, the classifier is fine-tuned with new data and used to estimate the entropy of the remaining data. The next informative batch is selected and also added to the informative set. Phases are repeated until the number of accumulated informative samples reaches a pre-set informative portion (IP). For example, $IP=0.3$ will select 30\% training samples as informative samples.}

\section{The Problem of Learning From Imbalanced Datasets}
\label{sec:problem}
In this section, a concise overview of the challenge of acquiring knowledge from imbalanced datasets is presented. Although the problem may apply to different machine learning methods, this study only focus on deep learning. 

\begin{figure}[t!]
	\includegraphics[width=\linewidth, trim=300 150 310 120,clip]{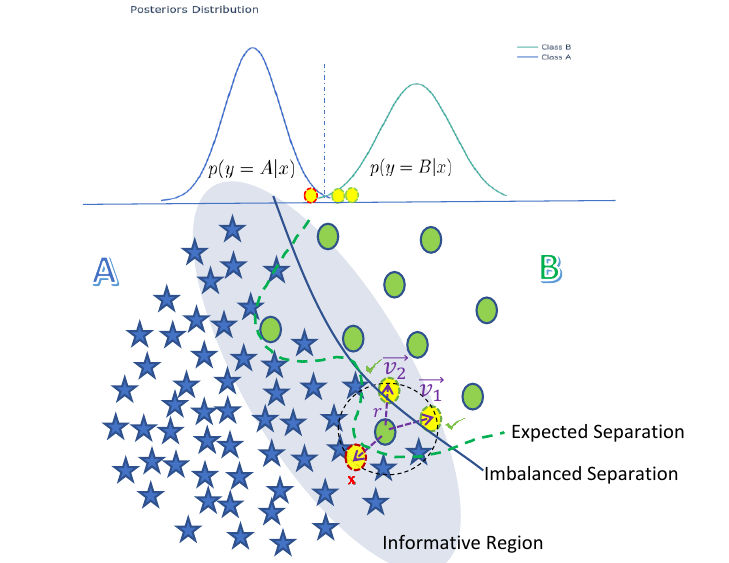}
	\caption{Learning from imbalanced datasets}
	\label{fig:problem}
\end{figure}

Figure \ref{fig:problem} illustrates our problem on binary classification. The imbalance in the informative region (light blue eclipse) could lead to classification errors. The dashed green line depicts the expected boundary, while the solid blue line is the model's boundary. Since the minority class lacks data in this region, the majority class will dominate the model even with a few noisy poorly-placed samples, which leads to a shift of the model's boundary. In contrast to the study by Ertekin \textit{et al.} \cite{ertekin_learning_2007}, which assumes the informative region is more balanced by nature and proposes a solution that only classifies over the informative samples, our assumption is different. This work contemplates the scenario where the informative region comprises extensively imbalanced data, which we believe is common in most real-world scenarios. The problem could be more severe in a more complex setting, such as high-dimensional and topologically complex data. Therefore, we proposed a technique to tackle the problem by oversampling the minority class in an informative manner. The detail of the technique will be described in Section \ref{sec:SIMPOR_method}.

\section{Methodology}
\label{sec:SIMPOR_method}
To alleviate the negative effects of data imbalance, we propose a comprehensive approach, \MethodnameLong{} (\Methodname), which aims to generate synthetic samples for minority classes. First, the informative region that contains informative samples is determined and balanced by creating surrounding synthetic neighbors for minority samples. The remaining region is then fully balanced by arbitrarily generating minority samples' neighbors. The remainder of this section provides further information about how our approach was developed.  

\subsection{Methodology Motivation}  
As Chazal and Michel mentioned in their work \cite{leroueil_compressibility_1996}, the natural way to highlight the global topological structure of the data is to connect data points' neighbors; our proposed method aligns with their observation by generating surrounding synthetic neighbors for minority samples to preserve data topology. Thus, our technique not only generates more data for minority class but also preserve the underlying topological structure of the entire data. 

Similar to \cite{ertekin_learning_2007} and \cite{aggarwal_active_2020}, we believe that informative samples play the most important role in the prediction success of both traditional machine learning models (e.g., SVM, Naive Bayes) and modern deep learning approaches (e.g., neural network). Thus, our technique finds these informative samples and focuses on augmenting minority data in this region. In this work, an entropy-based active learning strategy mentioned in \ref{sec:EAL} is applied to find the samples that contain more information to the model. This strategy is perhaps the most popular active learning technique and over-performs many other techniques on several datasets \cite{DAL}, \cite{7393573} \cite{settles_analysis_2008}.

\subsection{Generating minority synthetic data}  
\label{sec:solvingOptimization}
\Copy{1.1a}{
A synthetic neighbor $ x'$ and its label $y'$ can be created surrounding a minority sample $x$ by adding a small random vector $v$ to the sample, $x' = x + v$. Thus, $x'$ can be selected on the d-sphere's surface centered at $x$ with a radius of $|\vec{v}|$. For notation convenience, let $r = |\vec{v}|$ be the radius of the d-sphere. To enrich the synthetic samples, $r$ is sampled from a defined Gaussian distribution to generate a new synthetic sample distance each time. This section describes how the direction and distance of a synthetic sample are determined, which can also be represented via the direction and length of vector $\vec{v}$.

It is critical to generate synthetic data in the informative region because synthetic samples can unexpectedly jump across the decision boundary. This can be harmful to models as this might create outliers and reduce the model's performance. Therefore, we safely find vector $\vec{v}$ towards the minority class, such as $\vec{v}_0$ and $\vec{v}_1$ depicted in Figure \ref{fig:problem}.} Our technique is described via a binary classification scenario as follows. 

Let's consider a binary classification problem between majority class A and minority class B. 
From the Bayes' theorem, the posterior probabilities $p(y'=A|x')$ or $p(y'=B|x')$ can be used to present the probabilities that a synthetic sample $x'$ belongs to class A or class B, respectively. Let the two posterior probabilities be $f_0$ and $f_1$; they can be expressed as follows. 
\begin{align}
	\label{eq:posterior}
	p(y'=A|x') = \frac{p(x'|y'=A)\:p(A)}{p(x')} = f_0 \\
	p(y'=B|x') = \frac{p(x'|y'=B)\:p(B)}{p(x')} = f_1  
\end{align}

As mentioned earlier, each synthetic data $x'$ is generated so that it maximizes the probability of $x'$ belonging to the minority class $B$ and minimizes the chance $x'$ falling into the majority class $A$. Thus, a technique that maximizes the fractional posterior $f$ is proposed,   
\begin{align}
	\label{eq:fracpost}
	f &= f_1/f_0  \\
	&=\frac{p(x'|y'=B) \:p(B)}{p(x'|y'=A) \: p(A)}. \label{equ:f_ratio}
\end{align}

\textbf{\textit{Approximation of likelihoods in Equation \ref{equ:f_ratio}:}} A non-parametric kernel density estimates (KDE) is selected to approximate the likelihoods $p(x'|y'=A)$ and $p(x'|y'=B)$ as KDE is flexible and does not require specific assumptions about the data distribution. One can use a parametric statistical model such as Gaussian to approximate the likelihood; however, it oversimplifies the data and does not work effectively with topological complex data, especially in high dimensions. In addition, parametric models require an assumption about the distribution of data which is difficult in real-world problems since we usually do not have such information. On the other hand, KDE only needs a kernel working as a window sliding through the data. Among different commonly used kernels for KDE, we choose Gaussian Kernel as it is a powerful continuous kernel that would also eases the derivative computations for finding optima.
 
\Copy{PriorApproximation}{
\textbf{\textit{Approximation of priors in Equation \ref{equ:f_ratio}:}} Additionally, we estimate the prior probabilities of observing samples in class A ($p(A)$) and class B ($p(B)$) (in Equation \ref{equ:f_ratio}) by the widely-used Empirical Bayes Method \cite{empiricalBayes} to leverage the existing information from the original data. The estimates are denoted as $\widehat{p(A)}$ and $\widehat{p(B)}$ respectively.

\textbf{\textit{Equation \ref{equ:f_ratio} Approximation:} } Let $X_A$ and $X_B$ be the subsets of dataset $X$ which contain samples of class A and class B, $X_A=\{x: y=A  \}$ and $X_B=\{x: y=B  \}$. $N_A$ and $N_B$ are the numbers of samples in $X_A$ and $X_B$. $d$ is the number of data dimensions. $h$ presents the width parameter of the Gaussian kernel. The posterior ratio for each synthetic sample $x'$ then can be estimated as follows:

\begin{align}
	\label{eq:fracpost_estimation}
	f &= \frac{p(x'|y'=B)  \: p(B)}{p(x'|y'=A) \: p(A)} \\
	&\propto \frac{ \frac{1}{N_B h^d} \: \sum_{i=1}^{N_B}{ (2\pi)^{-\frac{d}{2}} \: e^{\frac{1}{2}{(\frac{x'-X_{B_i}}{h})^2} } } \: \widehat{p(B)} }
	{ \frac{1}{N_A h^d} \:  \sum_{j=1}^{N_A}{ (2\pi)^{-\frac{d}{2}} \: e^{ \frac{1}{2} {(\frac{x-X_{A_j}}{h})^2} } }\: \widehat{p(A)} }\\
	& \propto \frac{ \frac{1}{N_B h^d} \: \sum_{i=1}^{N_B}{ \: e^{\frac{1}{2}{(\frac{x'-X_{B_i}}{h})^2} } } \: \widehat{p(B)} }
	{ \frac{1}{N_A h^d} \:  \sum_{j=1}^{N_A}{  \: e^{ \frac{1}{2} {(\frac{x-X_{A_j}}{h})^2} } }\: \widehat{p(A)} }
	\label{equ:f}
\end{align} 
}

\textbf{\textit{Selecting bandwidth parameter $h$ for Gaussian kernel:} } The bandwidth is automatically selected for each dataset using the most common method, namely Scott's rule of thumb, proposed by Scott \cite{scott_2015}. With an attempt to minimize the mean integrated squared error, the parameter is estimated as $h = N^{(-\frac{1}{d+4})}$ where $N$, $d$ are the number of data points and the number of dimensions respectively. This study utilizes a scikitlearn python library for KDE, including bandwidth selection. The implementation detail can be found at \cite{skitlearnKDE}. 

\textbf{\textit{Finding synthetic samples surrounding a minority sample:}} To generate neighbors for each minority sample that maximizes Function \text{f} in Equation \ref{equ:f}, points on each $r$-radius sphere centered at a minority sample are considered synthetic instances. As a result, a vector $\vec{v}$ can be added to a minority sample for generating a new instance. \Copy{1.1b}{The relationship between a synthetic sample $x'$ and a minority sample can be described as follows,
\begin{align}
	\label{equ:vecV}
	\vec{x'} =  \vec{x} + \vec{v},
\end{align}
where the length of $\vec{v}$ is equal to $r$, and $r$ is sampled from a Gaussian distribution,
\begin{align}
     r \sim \mathcal{N}(0,\,(\alpha R)^{2}),
     \label{equ:r_dist}
\end{align}
where $\alpha R$ is the standard deviation of the Gaussian distribution and $ 0< \alpha <=1 $.
The range parameter $R$ is relatively small and computed as the average distance of a minority sample $x$ to its k-nearest neighbors. This will ensure that the generated sample will surround the minority sample. The Gaussian distribution with the mean of zero and the standard deviation $\alpha R$ controls the distance between the synthetic samples and the minority sample. The standard deviation is tuned from 0 to R by a coefficient $\alpha \in (0,1]$. The larger the $\alpha$ is, the farther synthetic data is placed from its original sample.} Consider a minority sample $x$ and its k-nearest neighbors in the Euclidean space, $R$ can be computed as follows:

\begin{align}
	R = \frac{1}{k}\sum\limits_{1}^{k} ||x-x_j ||,
\end{align}
where $||x-x_j||$ is the Euclidean distance between a minority sample $x$ and its $j$th neighbor. $k$ is a parameter indicating selected number of neighbors.

\begin{figure}[th]
	\includegraphics[width=\linewidth, trim=240 120 390 160,clip]{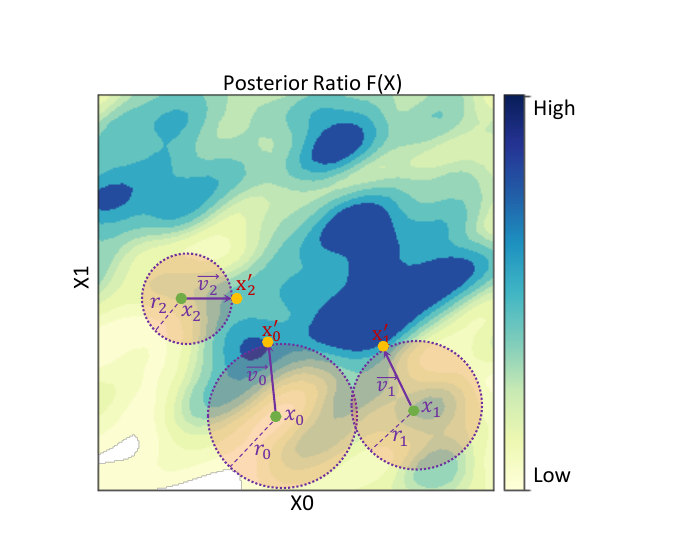}
	\caption{Demonstration on how \Methodname{} generates three synthetic samples $x'_0, x'_1, x'_2$, from three minority samples $x_0, x_1, x_2$, by maximizing the Posterior Ratio. }
	\label{fig:sphere_maxF}
\end{figure}

Figure \ref{fig:sphere_maxF} depicts a demonstration of finding 3 synthetic samples from 3 minority samples. In practice, one minority can be re-sampled to generate more than one synthetic samples. For a minority sample $x_0$, we find a synthetic sample $x_0'$ by maximizing the objective function $f(x_0'), x_0' \in X$ with a constraint that the Euclidean length of $\vec{v_0}$ equals to a radius $r_0$, $||\vec{v_0}|| = r_0$ or $||\vec{x_0'}-\vec{x_0}||=r_0$ (derived from Equation \ref{equ:vecV}).

The problem can be described as a constrained optimization problem. For each minority sample $x$, we find a synthetic sample $x'\in \mathbb{R}^d$ lying on the d-sphere centered at $x$ with radius $r$ and maximizing function in Equation \ref{equ:f},
\begin{align}
	\label{prob:optimazation}
	\max_{x'} {f(x')} \;\;\; \textrm{s.t.}\; ||\vec{x'} - \vec{x}||=r.
\end{align}

\textbf{\textit{Solving optimization problem in Equation \ref{prob:optimazation}:}} Interestingly, the problem in Equation\ref{prob:optimazation} can be solved numerically. Function $f(x)$ in Equation \ref{equ:f} is defined and continuous for $x' \in (-\infty, +\infty)$ because all of the exponential components (Gaussian kernels) are continuous and greater than zero. In addition, the constraint, $||\vec{x'} - \vec{x}||=r$, which contains all points on the sphere centered at $x$ with radius $r$ is a closed set (\cite{wikipedia_2021}). Thus, a maximum exits as proved in \cite{maximum_exist}. To enhance the diversity of synthetic data, either the global maximum or any local maximum can be accepted so that the synthetic samples will not simply go to the same direction.  

We solve the problem in Equation \ref{prob:optimazation} by using the Projected Gradient Ascent approach in which we iteratively update the parameter to go up the gradient of the objective function. A local maximum is found if the objective value cannot be increased by any local update. For simplification, we rewrite the problem in Equation \ref{prob:optimazation} by shifting the origin to the considered minority sample. The problem becomes finding the maximum of function $f(x')$, $x' \in \mathbb{R}^d$, constrained on a d-sphere, i.e., $||x'||=r$. Our solution can be described in Algorithm \ref{alg:optimization}. After shifting the coordinates system, we start by sampling a random point on the constraint sphere (line $1-2$). The gradient of the objective function at time $t$, $g_t(x'_t)$, is computed and projected onto the sphere tangent plane as $p_t$ (line $4-5$). It is then normalized and used for update a new $x'_{t+1}$ by rotating a small angle $lr*\theta$ (line $6-7$). The algorithm stops when the value of $f(x')$ is not increased by any update of $x'$. We finally shift to the original coordinates and return the latest $x'_t$.

\textbf{\textit{Avoiding synthesis of noise:}} To reduce the chance of misplacing synthetic samples on another class region because of noisy borderline and mislabeled minority samples, we set a policy for rejecting minority candidates which are selected for oversampling. The idea is to reject candidates surrounded mainly by other class samples. More specifically, we count the labels of the candidate's $k$-nearest neighbors and reject this candidate if there exists a class that its' number of samples is greater than the number of the minority samples. For example, the candidate is rejected when a class-A sample is selected for generating synthetic data, and its 5-nearest neighbors contain four class-B samples and one class-A sample. This is to avoid selecting mislabeled samples and noisy borderline samples for oversampling.        

\subsection{Algorithm}
Our strategy can be described in Algorithm \ref{alg:SIMPOR}. The algorithm takes an imbalanced dataset as its input and results in a balanced dataset which is a combination of the original dataset and synthetic samples. We first choose an active learning method $AL(\cdot)$ and find a subset of informative samples $S$ by leveraging entropy-based active learning (lines $1-2$). We then generate synthetic data to balance $S$. For each random sample $x_i^c$ in $S$ and belonging to minority class $c$, we randomly sample a small radius $r$ and find a synthetic sample that lies on the sphere centered at $x_i^c$ and maximizes the posterior ratio in Equation \ref{equ:f} (lines $3-11$). The process is repeated until the informative set $S$ is balanced. Similarly, the remaining region is balanced, which can be described in the pseudo-code from line $12$ to line $20$. The final output of the algorithm is a balanced dataset $D'$.


\section{Algorithm Time Complexity.}
\label{sec:implementation}

\Copy{timeComplexity}{

The costly part of \Methodname{} is that each synthetic sample requires computing a kernel density estimation of the entire dataset. Elaborately, let $n$ be the number of samples of the dataset. In the worst case, the numbers of samples of minority and majority class are $N_B = 1$ and $N_A = n-1$, respectively. We need to generate $n-2$ synthetic samples to balance the dataset completely. Since each generated sample must loop through the entire dataset of size $n$ to estimate the density, the algorithm complexity is $O(n^2)$. 

Although generating synthetic data is only a one-time process, and this does not affect the classification efficiency in the testing phase, we still try to alleviate its weakness by providing parallelized implementations to reduce the time complexity to $O(n)$. Specifically, each exponential component in Equation \ref{equ:f} is computed parallelly, utilizing GPU or CPU threads. Ellaborately, Equation \ref{equ:f} can be rewritten as $N_B$ components of $e^{\frac{1}{2} (\frac {x - X_{B_i}}{h})^2}$ and $N_A$ components of $e^{\frac{1}{2} (\frac {x - X_{A_i}}{h})^2}$. Fortunately, they are all independent and can be processed parallelly. Thus, with a sufficient hardware resource, the consumption time for the kernel density estimation of each synthetic data point is then reduced by $N_A+N_B=n$ times, which significantly simplifies the complexity to $O(n)$.

}

\section{Experiments}
\label{sec:experiments}
\Copy{sow}{
In this section, we explore the techniques via binary classification problems on an artificial dataset (i.e., Moon) and 41 real-world datasets i.e., Knowledge Extraction based on Evolutionary Learning (KEEL), University of California Irvine (UCI), Credit Card Fraud, with a diversity of imbalance ratios and different numbers of features. Samples in Moon have two features, while other datasets contain various numbers of features and imbalance ratios.} Dataset details are described in Table \ref{tab:dataDecription}.  
The implementation steps to balance datasets follow Algorithm \ref{alg:SIMPOR}. To evaluate our proposed balancing technique, we compare the classification performance to different widely-used and state-of-the-art techniques. More specifically, We compare \Methodname{} to SMOTE \cite{chawla_smote:_2002}, Borderline-SMOTE \cite{bordersmote},  ADASYN \cite{ADASYN}, DeepSMOTE \cite{deepsmote}, Gaussian Distribution Based Oversampling (GDO) \cite{bib:GDO}, SVMCS \cite{cssvm}, EE \cite{EE}. To evaluate the classifications performance for skewed datasets, we measure widely-used metrics, i.e., F1-score and Area Under The Curve (AUC).

\subsection{Implementation Detail}
This section describes the general settings and implementation details for the experimental techniques. Our implementation code is publicly available on Github \footnote{\url{https://github.com/nsh135/_SIMPOR_}}.

\subsubsection{\Methodname{} settings}

In order to find the informative subset, we leverage entropy-based active learning. We first utilize a neural network model playing a role as a classifier to find high-entropy samples (Note that the classifier for finding the informative subset differs from the classifiers for the final classification evaluation after all balancing techniques are applied to the data). The detailed steps are introduced in Section \ref{sec:EAL}. The model contains two fully connected hidden layers with \textit{relu} activation functions and 10 neurons in each layer. The output layer applies the soft-max activation function. The model is trained in a maximum of 300 epochs with an early stop option when the loss is not significantly improved after updating weights. The model is trained firstly on a random set of three samples each class (six samples two classes). This model is then used to estimate entropy scores for the remaining data. We then select next 20  highest entropy samples ($k$=20) for the next informative data batch. This batch is concatenated to the initial batch for updating the classifiers and accumulated to the informative set. The steps are repeated until the informative set reaches desire informative portion (IP). In these experiments, we set IP=0.3 corresponding to 30 percent of the training size selected for the informative set.

To solve the optimization problem in Equation \ref{prob:optimazation} for finding optima (this differs from the classification optimization for the evaluation) introduced in Section \ref{sec:solvingOptimization}, we use a gradient ascent method with the gradient rate of $1e-5$ and the maximum iteration of 300.   

\subsubsection{Evaluation Classification settings}
Considering each imbalanced dataset as a classification problem, we use the classification testing performance for the technique comparison. Each dataset is randomly split into two parts, 80\% for training and 20\% for testing. The classifiers are trained on training sets after applying the techniques. The results are reported on the raw testing sets (There isn't any technique applied on the testing sets; thus, they are also possibly class imbalanced). We use F1-score and AUC for the evaluation metrics as they are suitable and widely used to evaluate imbalanced data. Reported testing results for each dataset are the averages of 5 experimental trials.

The classifiers are constructed by neural networks with the input and output sizes corresponding to the number of datasets' features and unique labels. We use the same classifier structure (number of hidden layers, number of neurons in each layer, learning rate, optimizer) for all compared datasets. The detail of neural network implementation is described in Table \ref{tab:model_setting}. 
For baseline technique settings, we follow the experimental parameter sets in \cite{bib:GDO} as we share very similar datasets and comparison techniques. For DeepSMOTE settings, the DCGAN input and output sizes are modified to adapt with each dataset, while other settings is taken from the initial parameter set in \cite{deepsmote}.

\begin{table}[htbp!]
	\centering
	\caption{Classification models' setting for each dataset.}
	\label{tab:model_setting}
	\resizebox{0.9\columnwidth}{!}{%
	
\begin{tabular}{lp{29.57em}}
	\toprule
	Method & \multicolumn{1}{l}{Parameter} \\
	\midrule
	SIMPOR & \multicolumn{1}{l}{k\_neighbors=5, r\_distribtuion=Gaussian, IP=0.3} \\
	GDO   & \multicolumn{1}{l}{k\_neighbors=5, d=1} \\
	SMOTE & \multicolumn{1}{l}{k\_neighbors=5, sampling\_strategy=`auto',random\_state=None} \\
	BL-SMOTE & \multicolumn{1}{l}{k\_neighbors=5, sampling\_strategy=`auto', random\_state=None} \\
	ADASYN & \multicolumn{1}{l}{k\_neighbors=5, sampling\_strategy=`auto', random\_state=None} \\
	EE    & \multicolumn{1}{l}{\#estimators=10, Estimater=AdaBoostClassifier} \\
	DeepSMOTE   & \multicolumn{1}{l}{Sigma=1, Lambda=0.1} \\
	\midrule
	Classifier & \multicolumn{1}{l}{Parameter} \\
	\midrule
	Architecture & \multicolumn{1}{l}{neuron/layer=100, \#layers=3} \\
	Optimization & optimizer=`adam',  epochs=200, batch\_size=32, learning\_rate=0.1, reduce\_lr\_loss(factor=0.9,epsilon=1e-4,patience=5) \\
	\bottomrule
\end{tabular}%

	}
\end{table}%

\subsection{\Methodname{} on artificial Moon dataset}

\begin{figure}[h!]	
	\includegraphics[width=0.9\linewidth]{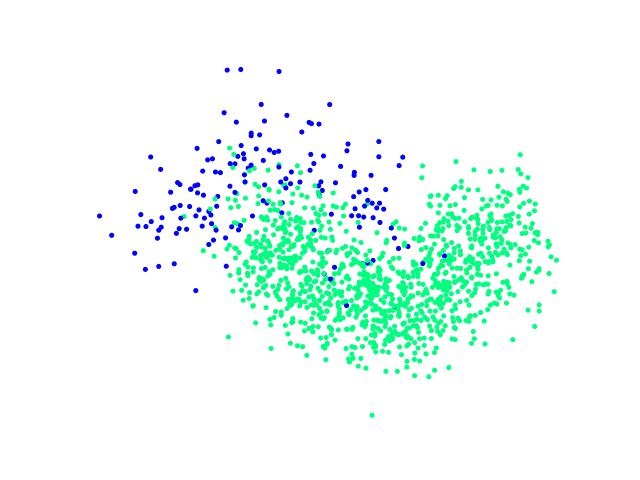}
	\caption{Artificial class imbalanced Moon dataset with IR of 7:1.}
	\label{fig:raw_moon}
\end{figure}

\begin{figure*}[th]
	\centering
	\begin{subfigure}[]{0.3\linewidth}
		\includegraphics[width=\linewidth]{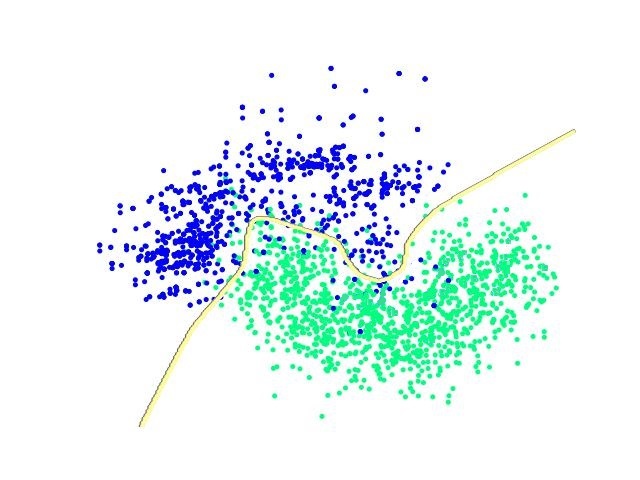}
		\caption{\Methodname{}.}
		\label{fig:simpor_moon}
	\end{subfigure}
	\hspace{0.1em}%
	\begin{subfigure}[]{0.3\linewidth}
		\includegraphics[width=\linewidth]{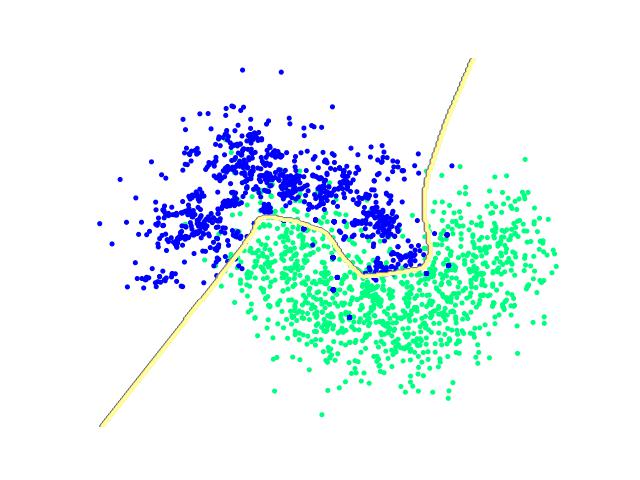}
		\caption{GDO.}
		\label{fig:gdo_moon}
	\end{subfigure}
	\hspace{0.1em}%
	\begin{subfigure}[]{0.3\linewidth}
		\includegraphics[width=\linewidth]{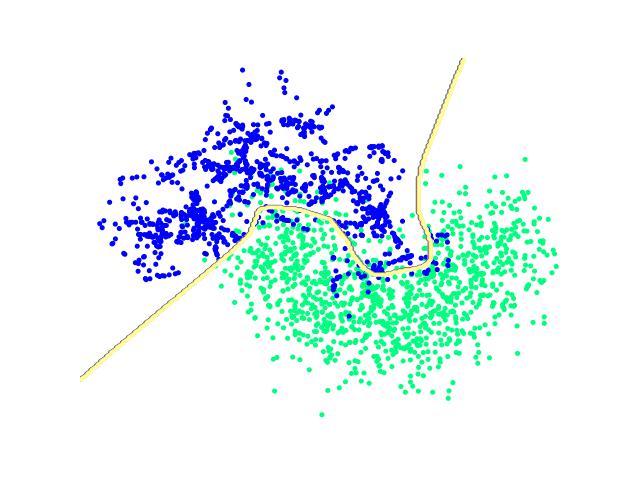}
		\caption{SMOTE.}
		\label{fig:smote_moon}
	\end{subfigure}
	\\
	\begin{subfigure}[]{0.3\linewidth}
		\includegraphics[width=\linewidth]{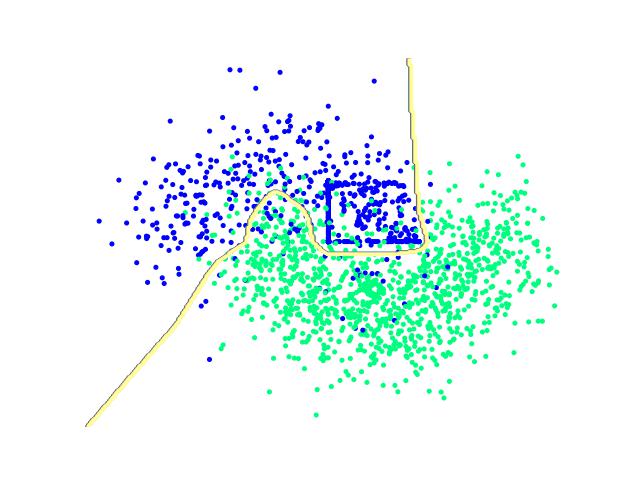}
		\caption{ DeepSMOTE.}
		\label{fig:deepsmote_moon}
	\end{subfigure}
	\hspace{0.1em}%
	\begin{subfigure}[]{0.3\linewidth}
		\includegraphics[width=\linewidth]{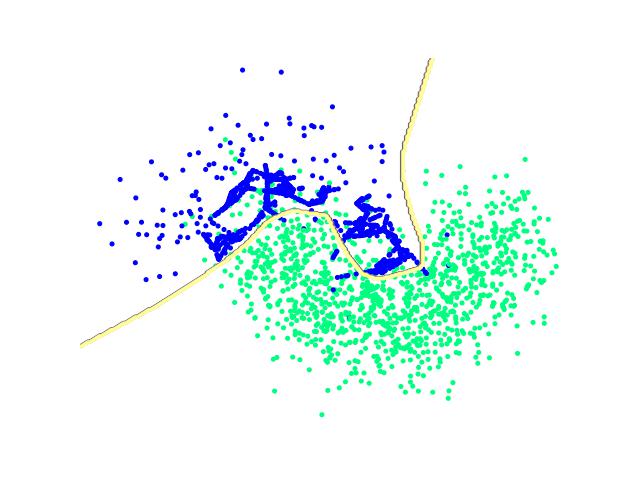}
		\caption{BorderlineSMOTE.}
		\label{fig:border_smote_moon}
	\end{subfigure}
	\hspace{0.1em}%
	\begin{subfigure}[]{0.3\linewidth}
		\includegraphics[width=\linewidth]{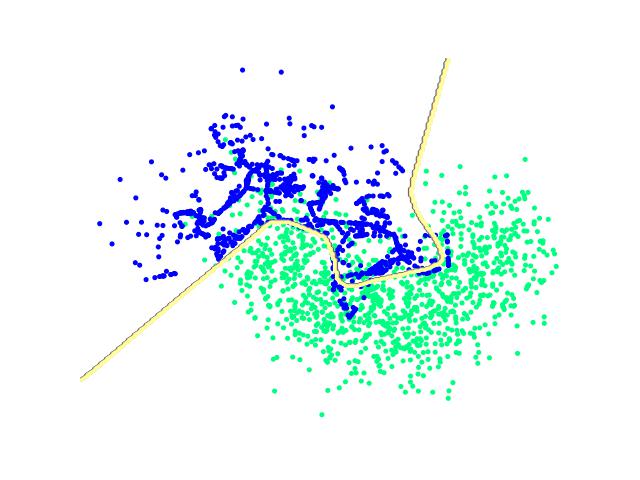}
		\caption{ADASYN. }
		\label{fig:adasyn_moon}
	\end{subfigure}
	
	\caption{Data plot and model's decision boundary visualization for Moon Dataset over different techniques.}
	\label{fig:MoonResults}
\end{figure*}

\Copy{moonGeneration}{
We implement techniques on an artificial 2-dimension dataset for demonstration purposes. We first generate the balanced synthetic MOON dataset using python library \textit{sklearn.datasets.make\_moons}. The generated MOON contains 3000 samples labeled in two classes, and each instance has two numerical features with values ranging from 0 to 1. We then make the dataset artificially imbalanced with an Imbalance Ratio of 7:1 by randomly removing 1285 samples from one class.} As a result, the training dataset becomes imbalanced, as visualized in Figure \ref{fig:raw_moon}.

Figure \ref{fig:MoonResults} captures the classification for different techniques. We also visualize the model decision boundaries to provide additional information on how the classification models are affected. We use a fully connected neural network described in Table \ref{tab:model_setting} to classify the data.

\begin{table}[htbp!]
	\centering
	\caption{Classification Result on Moon Dataset.}
	\resizebox{\columnwidth}{!}{%
		
	\begin{tabular}{crcccccc}
		Metric &       & SIMPOR & SMOTE & BL-SMOTE & DeepSMOTE   & ADASYN & GDO \bigstrut[b]\\
		\hline
		F1-score &       & 0.883 & 0.824 & 0.827 & 0.842 & 0.785 & 0.817 \bigstrut[t]\\
		AUC   &       & 0.961 & 0.957 & 0.955 & 0.959 & 0.955 & 0.959 \\
	\end{tabular}%

}
	\label{tab:MoonPerformance}
\end{table}%

\subsubsection{Results and Discussion}
From the visualization shown in Figure \ref{fig:MoonResults} and the classification performance results in Table \ref{tab:MoonPerformance}, it is clear that \Methodname{} performs better than others by up to 10\% on F1-score and 1.1\% on AUC. We can see that DeepSMOTE (DeepSM) creates dense squared noise and pushes the decision boundary to the majority class. Due to the fact that SMOTE-based methods does not take the informative region into account, unbalanced data in this area lead to a severe error in decision boundary. In Figures \ref{fig:adasyn_moon} and \ref{fig:border_smote_moon}, BorderlineSMOTE (BL-SMOTE) and ADASYN focus on the area near the model's decision boundary, but they inherit a drawback from SMOTE; any noise or mislabeled samples can, unfortunately, create very dense bridges crossing the expected border and lead to decision errors. Figure \ref{fig:gdo_moon} shows that GDO also generates local gaussian groups of samples near the boder and thus create errors. This phenomenon might cause by a few mis-labeled sample points. In contrast, by generating neighbors of minority samples in the direction towards the minority class and balancing the informative region, \Methodname{} (Figure \ref{fig:simpor_moon}) helps the classifier to make a better decision with a solid smooth decision boundary. Poorly-placed synthetic samples are significantly less than that of others.

\subsection{\Methodname{} on forty-one real datasets}
\label{subsec:41Datasets}
\Copy{datasetDetail}{
In this section, we compare the proposed technique on 41 real two-class datasets with a variable number of features and Imbalance Ratios, i.e., KEEL datasets \cite{ KEEL_detail,KEEL_dataset}, UCI datasets fetched from Sklearn tool \cite{imbalancedlearnFetch_datasetsx2014, uci_imbalance_dataset} and Credit Card Fraud \cite{kaggleCreditCard} dataset. Since the original Credit Card Fraud contains a large number of banking normal and fraud transaction samples (284,807) which significantly reduces our experimental efficiency, we reduced the dataset size by randomly removing normal class transactions to reach an imbalance ratio of 3.0. Other datasets are kept as their original versions after removing bad samples (containing Null values). The datasets are described in Table \ref{tab:dataDecription}. 
}
\subsubsection{Classification results}
\label{sec:classificationResult}

\begin{figure}[h!]
	\centering
	\includegraphics[width=\linewidth ]{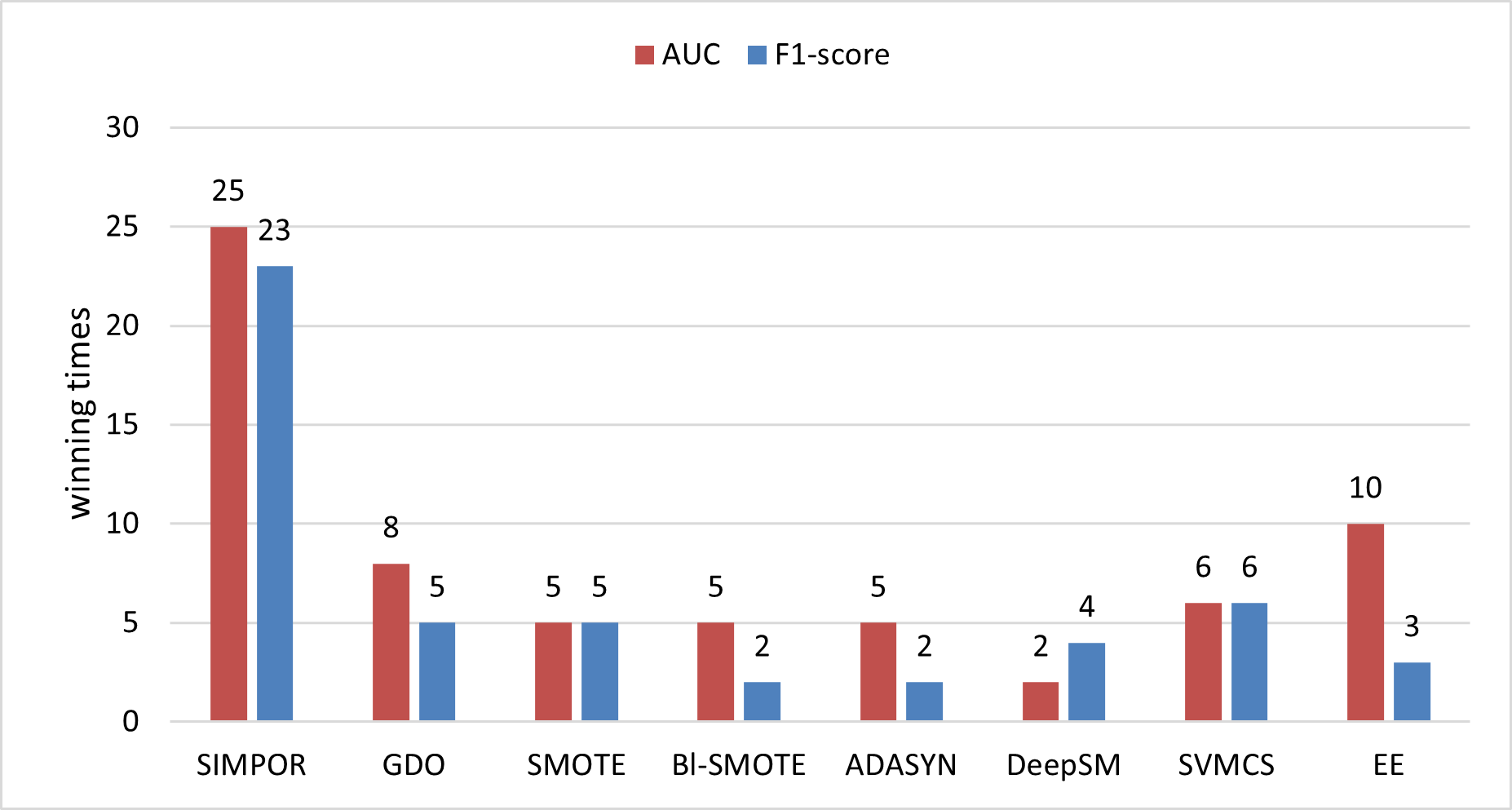}
	\caption{Winning times over 41 datasets.}
	\label{fig:winingTimes}
\end{figure}

Table \ref{tab:F1AllDatasets}, \ref{tab:AUCAllDatasets}, \ref{tab:Precision} and \ref{tab:Recall} show the classification F1-score, AUC, Precision, and Recall results, respectively. The highest scores for each dataset are highlighted in bold style. 
\Copy{winningTime}{We also provide the summary of the F1 and AUC scores by ``winning times" scores. We count the number of datasets for which a technique achieves the highest scores among the compared techniques and name this number ``winning times". For convention, if more than two techniques share the same highest score, the winning times will be increased for each technique. Figure \ref{fig:winingTimes} shows a summary of winning times.} 

As we can see from the table, the proposed technique outperforms others on both evaluation metrics, F1-score and AUC. More specifically, \Methodname{} hits 23 F1-score winning times and 25 AUC winning times. Its number of F1-score winning times at 23 four times better than the second winner (SVMCS) at 6, and its AUC winning times at 25 doubles the second AUC winners (EE) at 10.

\subsection{Statistical Test.}
	\label{sec:wilcoxon}

To further evaluate the effectiveness of the technique, we also performed a Wilcoxon Signed Rank Test \cite{wilcoxon} on the 41 dataset results (F1 score and AUC). Wilcoxon hypothesis test is relevant to our study as it is a non-parametric statistical test and does not require a specific distribution assumption for the results. On the other hand, 41 data points (corresponding to 41 datasets results) are sufficient to support this test. Our null hypothesis is that the difference between the proposed technique results and those of the other technique is insignificant. Wilcoxon signed-rank test outputs are computed over the 41 dataset results and return a p-value for each technique pair. We then compare the p-value with the significant value $\alpha = 0.05$. Suppose the p-value is smaller than $\alpha$. In that case, the evidence is sufficient to reject the hypothesis, which means the proposed technique does make a significant difference from the others, and vice versa. Table \ref{tab:wilcoxonTest} shows the Wilcoxon p-value results.

	\begin{table}[htbp]
		\centering
		\caption{Wilcoxon Signed Rank Hypothesis Test results.}
		
		\begin{tabular}{lcc}
			\toprule
			& \multicolumn{2}{c}{p-value} \\
			\midrule
			SIMPOR vs. & \multicolumn{1}{l|}{F1-score} & \multicolumn{1}{l}{AUC} \\
			\midrule
			GDO   & 1.82E-03 & 1.93E-03 \\
			SMOTE & 2.66E-03 & 6.64E-05 \\
			BL\_SMOTE & 4.22E-03 & 1.63E-04 \\
			ADASYN & 2.89E-03 & 6.13E-04 \\
			DeepSM   & 6.40E-04 & 2.57E-04 \\
			SVMCS & 2.74E-03 & 3.40E-02 \\
			EE    & 1.99E-03 & 2.17E-02 \\
			\bottomrule
		\end{tabular}%
		\label{tab:wilcoxonTest}%
	\end{table}%

As we can see from Table \ref{tab:wilcoxonTest}, the p-values are all smaller than the critical value of 0.05. Thus, the null hypothesis can be rejected as the supporting evidence is sufficient. In other words, the statistical result shows that the proposed technique makes a significant improvement compared to others.

\subsection{Processing Time.}
\label{sec:processingTime}
Data processing times for oversampling-based approaches on 41 datasets are compared to provide a more comprehensive comparison. We don't compare them to the other approaches, i.e., cost-sensitive learning and ensemble learning, because they only need negligible data processing time as they focus on classifiers other than improving the data. The processing time was recorded from our machine, which uses an Intel i7 32-thread processor and two NVIDIA 3090 Ti GPUs. 
Table \ref{tab:ProcessingTime} shows the recorded processing time over 41 datasets. Overall, our technique takes longer than others as we have to compute the kernel estimation for each data point, as mentioned in Section \ref{sec:implementation}. Similarly, DeepSMOTE generally suffers high time consuming cost because it heavily relies on underlying heuristic methods. In other words, the proposed technique is slower, but it provides better F1 and AUC scores than others.

\begin{table}[htbp]
	\centering
	\caption{Processing time (in seconds) over 41 datasets.}
	\resizebox{0.97\columnwidth}{!}{%
		
		\begin{tabular}{lcccccc}
			\toprule
			& SIMPOR & GDO   & SMOTE & BL-SMOTE & ADASYN & DeepSM \\
			\midrule
			glass1 & 0.1147 & 0.0576 & 0.0020 & 0.0033 & 0.0032 & 0.8587 \\
			wisconsin & 2.0805 & 0.1769 & 0.0024 & 0.0044 & 0.0046 & 1.2004 \\
			pima  & 0.2032 & 0.2066 & 0.0025 & 0.0049 & 0.0050 & 1.2297 \\
			glass0 & 0.2157 & 0.0553 & 0.0023 & 0.0035 & 0.0036 & 0.8601 \\
			yeast1 & 0.2457 & 0.4749 & 0.0035 & 0.0108 & 0.0104 & 1.4846 \\
			haberman & 0.0517 & 0.1560 & 0.0022 & 0.0033 & 0.0036 & 0.9246 \\
			vehicle1 & 0.4365 & 0.1237 & 0.0025 & 0.0059 & 0.0059 & 1.4147 \\
			vehicle2 & 6.2913 & 0.1512 & 0.0029 & 0.0053 & 0.0061 & 1.3976 \\
			vehicle3 & 0.2821 & 0.1237 & 0.0024 & 0.0060 & 0.0061 & 1.3487 \\
			creditcard & 2.1200 & 0.3783 & 0.0087 & 0.0184 & 0.0182 & 1.7980 \\
			glass-0-1-2-3\_vs\_4-5-6 & 0.3376 & 0.0459 & 0.0023 & 0.0035 & 0.0035 & 0.8407 \\
			vehicle0 & 7.3645 & 0.1198 & 0.0024 & 0.0054 & 0.0058 & 1.2953 \\
			ecoli1 & 0.0418 & 0.0337 & 0.0010 & 0.0018 & 0.0017 & 0.9310 \\
			new-thyroid1 & 0.5352 & 0.0304 & 0.0015 & 0.0024 & 0.0024 & 0.8590 \\
			new-thyroid2 & 0.3881 & 0.0359 & 0.0025 & 0.0033 & 0.0031 & 0.8747 \\
			ecoli2 & 0.2516 & 0.0266 & 0.0011 & 0.0017 & 0.0016 & 0.9733 \\
			glass6 & 0.3196 & 0.0268 & 0.0014 & 0.0025 & 0.0023 & 1.0744 \\
			yeast3 & 0.1374 & 0.2422 & 0.0023 & 0.0060 & 0.0059 & 1.6699 \\
			ecoli3 & 0.0658 & 0.0378 & 0.0015 & 0.0025 & 0.0024 & 0.9647 \\
			page-blocks0 & 7.9654 & 2.0918 & 0.0045 & 0.0143 & 0.0138 & 3.6029 \\
			yeast-2\_vs\_4 & 2.4310 & 0.0624 & 0.0017 & 0.0028 & 0.0028 & 1.0286 \\
			yeast-0-5-6-7-9\_vs\_4 & 0.0868 & 0.0632 & 0.0016 & 0.0029 & 0.0027 & 0.9809 \\
			vowel0 & 4.7675 & 0.1312 & 0.0018 & 0.0039 & 0.0037 & 1.2410 \\
			glass-0-1-6\_vs\_2 & 0.0482 & 0.0207 & 0.0013 & 0.0023 & 0.0022 & 0.9133 \\
			glass2 & 0.0501 & 0.0227 & 0.0013 & 0.0024 & 0.0024 & 0.8855 \\
			yeast-1\_vs\_7 & 0.4697 & 0.0420 & 0.0017 & 0.0026 & 0.0026 & 1.0355 \\
			glass4 & 0.1141 & 0.0197 & 0.0012 & 0.0024 & 0.0023 & 0.9469 \\
			ecoli4 & 0.1087 & 0.0310 & 0.0015 & 0.0024 & 0.0024 & 0.9393 \\
			page-blocks-1-3\_vs\_4 & 1.8742 & 0.0445 & 0.0015 & 0.0027 & 0.0026 & 0.9992 \\
			abalone9-18 & 2.9722 & 0.0716 & 0.0015 & 0.0028 & 0.0026 & 1.2095 \\
			yeast-1-4-5-8\_vs\_7 & 0.0881 & 0.0673 & 0.0017 & 0.0031 & 0.0028 & 1.0803 \\
			glass5 & 0.2815 & 0.0241 & 0.0017 & 0.0033 & 0.0036 & 0.8550 \\
			yeast-2\_vs\_8 & 0.1239 & 0.0441 & 0.0016 & 0.0027 & 0.0028 & 0.9849 \\
			car\_eval\_4 & 0.4381 & 0.1746 & 0.0026 & 0.0066 & 0.0049 & 1.6616 \\
			wine\_quality & 0.1622 & 0.8587 & 0.0030 & 0.0144 & 0.0137 & 3.3128 \\
			yeast\_me2 & 0.1060 & 0.1379 & 0.0018 & 0.0042 & 0.0039 & 1.7350 \\
			yeast4 & 0.1083 & 0.1386 & 0.0018 & 0.0041 & 0.0039 & 1.6314 \\
			yeast-1-2-8-9\_vs\_7 & 0.0924 & 0.0757 & 0.0017 & 0.0031 & 0.0030 & 1.3069 \\
			yeast5 & 0.1188 & 0.1312 & 0.0019 & 0.0037 & 0.0040 & 1.6181 \\
			yeast6 & 0.0613 & 0.1419 & 0.0018 & 0.0037 & 0.0036 & 1.6382 \\
			abalone19 & 0.0890 & 0.3161 & 0.0022 & 0.0053 & 0.0054 & 3.0168 \\
			\bottomrule
		\end{tabular}%
		
	}
	\label{tab:ProcessingTime}%
\end{table}%

\section {Conclusion}
\label{sec:conclusion}
A data balancing technique by oversampling the minority class is proposed. The technique aims at balancing datasets and preventing the creation of noise in data by directing the synthetic samples toward the minority class. Our experiment results show that the proposed technique outperforms other experimental techniques over 41 real-world datasets. For future work, we would like to investigate the class imbalance for image data type and enhance our approach to adapt to image datasets.   

\section*{Acknowledgement}

Efforts sponsored in whole or in part by United States Special Operations Command (USSOCOM), under Partnership Intermediary Agreement No. H92222-15-3-0001-01. The U.S. Government is authorized to reproduce and distribute reprints for Government purposes notwithstanding any copyright notation thereon.  
{\footnote{ The views and conclusions contained herein are those of the authors and should not be interpreted as necessarily representing the official policies or endorsements, either expressed or implied, of the United States Special Operations Command.} }

\bibliographystyle{IEEEtran}
\bibliography{citation}

\begin{IEEEbiography}[{\includegraphics[width=0.9\linewidth,clip,keepaspectratio]{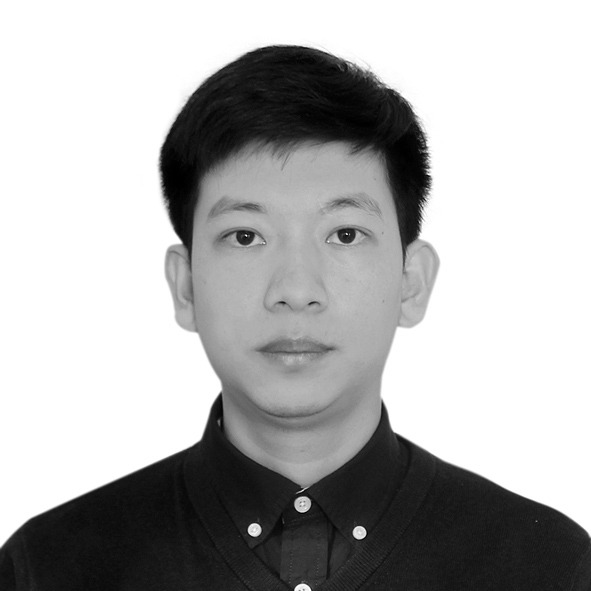}}]{Hung Nguyen}
	received the M.Sc. degree and currently pursuing his Ph.D. degree in Department of Electrical Engineering, University of South Florida, FL, USA. His current research interests include machine learning, artificial intelligence, federated learning, cyber security, privacy enhancing technologies. Hung is a member of IEEE.
\end{IEEEbiography}

%
%

\newpage
\appendices

\section{Agorithms}
\label{apx.Agorithms}

\begin{algorithm}[ht]
	\caption{Sphere-Constrained Gradient Ascent for Finding Maximum}
	
	\begin{flushleft}
		\textbf{Input}: A minority sample $x_0$, objective function $f(x,X)$\\
		\textbf{Parameter}: \\
		$r$ : The radius of the sphere centered at $x_0$  \\\
		$\theta$ : Sample space $\theta \in [0,2\pi]$ \\
		$lr$ : Gradient ascent learning rate\\
		
		\textbf{Output}: An local maximum $x'$\\
		\begin{algorithmic}[1]
			\STATE Shift the Origin to $x_0$
			\STATE Randomly initiate $x'_t$ on the sphere with radius $r$	
			\WHILE {converge condition}
			\STATE Compute the gradient at $x'_t$\\
			$g_t(x'_t) = \nabla f(x'_t)$
			\STATE Project the gradient onto the sphere tangent plane\\
			$p_t = g_t - (g_t \cdot x'_t) x_t$
			\STATE Normalize projected vector\\
			$p_t = p_t/ ||p_t||$
			\STATE Update $x'$ on the constrained sphere \\
			$x'_{t+1} = x'_t cos(lr*\theta) + p_t sin (lr*\theta)$ 			
			\ENDWHILE
			\STATE Shift back to the Origin
			\RETURN $x'_t$
		\end{algorithmic}
	\end{flushleft}
	\label{alg:optimization}
\end{algorithm}

\begin{algorithm}[ht]
	\caption{\Methodname}
	
	\begin{flushleft}
		\textbf{Input}: Original Imbalance Dataset $D$ including data $X$ and labels $y$.\\
		\textbf{Parameter}: 
		$MA$ is the majority class, $MI$ is a set of other classes.\\
		$k$: Number of  neighbors of the considered sample which determines the maximum range of the sample to its synthetic samples.\\ 
		$\alpha$: preset radius coefficient
		$Count(c, P)$ : A function to count class $c$ sample number in population $P$.\\
		$G(x_0,f,r)$ : Algorithm \ref{alg:optimization}, which returns a synthetic sample on sphere centered at $x_0$ with radius $r$ and maximize Equation \ref{equ:f}.  \\
		\textbf{Output}: Balanced Dataset $D'$ including $\{X',y'\}$
		\begin{algorithmic}[1]
			\STATE Select an Active Learning Algorithm $AL()$
			\STATE Query a subset of informative samples $S \in D$  using $AL$:
			$s  \leftarrow AL(D) $	\\
			
			\COMMENT{Balance the informative region}
			\FOR {$c \in MI$}
			\WHILE {$Count(c, S) \leq Count(MA, S)$ }
			\STATE Select a random $x_i^c \in S$
			\STATE Reject and reselect $x_i^c$ if its label is dominated among k-nearest labels
			\STATE Compute maximum range $R$ based on k-nearest neighbors
			\STATE Randomly sample a radius $r \sim \mathcal{N}(0,\alpha R)$
			\STATE Generate a synthetic neighbor $x'$ from $x_i^c$:
			$x'=G(x_i^c,f,r)$ 
			\STATE Append $x'$ to $D'$
			\ENDWHILE
			\ENDFOR
			
			\COMMENT{Balance the remaining region}
			\FOR {$c$ in $MI$}
			\WHILE { $Count(c, D') \leq Count(MA, D')$ }	
			\STATE Select a random $x_j^c \in \{X-S\}$
			\STATE Compute maximum range $R$ based on $k$
			\STATE Randomly sample a radius $r \sim \mathcal{N}(0,\alpha R)$
			\STATE Generate a synthetic neighbor $x'$ of $x_j^c$ 
			\STATE Append $x'$ to $D'$
			\ENDWHILE
			\ENDFOR
			\RETURN 
		\end{algorithmic}
	\end{flushleft}
	\label{alg:SIMPOR}
\end{algorithm}
\newpage

\section{Table Results on Different Metrics}
\label{apx.results}

\begin{table}[!htbp]
	\centering
	\caption{F1-score over different datasets.}
	\resizebox{0.97\columnwidth}{!}{%
		
		\begin{tabular}{lrrrrrrrr}
			\toprule
			& \multicolumn{1}{c}{SIMPOR} & \multicolumn{1}{c}{GDO} & \multicolumn{1}{c}{SMOTE} & \multicolumn{1}{c}{BL-SMOTE} & \multicolumn{1}{c}{ADASYN} & \multicolumn{1}{c}{DeepSM} & \multicolumn{1}{c}{SVMCS} & \multicolumn{1}{c}{EE} \\
			\midrule
			glass1 & 0.729 & \textbf{0.741} & 0.707 & 0.729 & 0.729 & 0.706 & 0.719 & 0.705 \\
			wisconsin & \textbf{0.962} & 0.959 & 0.953 & 0.958 & 0.956 & 0.960 & 0.958 & 0.957 \\
			pima  & \textbf{0.777} & 0.699 & 0.714 & 0.720 & 0.700 & 0.721 & 0.742 & 0.731 \\
			glass0 & \textbf{0.840} & 0.799 & 0.804 & 0.795 & 0.806 & 0.813 & 0.835 & 0.811 \\
			yeast1 & \textbf{0.715} & 0.676 & 0.675 & 0.685 & 0.672 & 0.673 & 0.685 & 0.678 \\
			haberman & \textbf{0.601} & 0.599 & 0.589 & 0.587 & 0.580 & 0.587 & 0.586 & 0.584 \\
			vehicle1 & \textbf{0.824} & 0.815 & 0.807 & 0.796 & 0.817 & 0.785 & 0.784 & 0.808 \\
			vehicle2 & \textbf{0.987} & 0.967 & 0.977 & 0.978 & 0.981 & 0.954 & 0.976 & 0.981 \\
			vehicle3 & \textbf{0.821} & 0.766 & 0.785 & 0.792 & 0.806 & 0.780 & 0.782 & 0.785 \\
			creditcard & \textbf{0.954} & 0.935 & 0.946 & 0.944 & 0.943 & 0.947 & 0.907 & 0.939 \\
			glass-0-1-2-3\_vs\_4-5-6 & 0.850 & 0.923 & 0.918 & 0.912 & 0.915 & \textbf{0.929} & 0.907 & 0.905 \\
			vehicle0 & 0.933 & 0.956 & \textbf{0.970} & 0.965 & 0.965 & 0.952 & \textbf{0.970} & 0.969 \\
			ecoli1 & 0.831 & 0.822 & 0.838 & 0.818 & 0.815 & \textbf{0.853} & 0.824 & 0.827 \\
			new-thyroid1 & 0.970 & \textbf{0.979} & 0.946 & 0.953 & 0.953 & 0.902 & 0.946 & 0.946 \\
			new-thyroid2 & 0.962 & \textbf{0.982} & 0.938 & 0.938 & 0.938 & 0.872 & 0.930 & 0.930 \\
			ecoli2 & \textbf{0.922} & 0.880 & 0.905 & 0.864 & 0.884 & 0.887 & 0.909 & 0.914 \\
			glass6 & \textbf{0.952} & 0.899 & 0.875 & 0.880 & 0.869 & 0.864 & 0.880 & 0.880 \\
			yeast3 & 0.862 & 0.818 & 0.842 & 0.836 & 0.829 & 0.831 & 0.867 & \textbf{0.879} \\
			ecoli3 & 0.806 & 0.791 & 0.790 & 0.792 & 0.792 & \textbf{0.829} & 0.827 & 0.824 \\
			page-blocks0 & \textbf{0.926} & 0.904 & 0.909 & 0.900 & 0.900 & 0.913 & 0.919 & 0.912 \\
			yeast-2\_vs\_4 & 0.883 & 0.875 & \textbf{0.893} & 0.844 & 0.866 & 0.817 & 0.807 & 0.772 \\
			yeast-0-5-6-7-9\_vs\_4 & \textbf{0.824} & 0.752 & 0.754 & 0.781 & 0.758 & 0.747 & 0.813 & 0.805 \\
			vowel0 & \textbf{1.000} & \textbf{1.000} & \textbf{1.000} & \textbf{1.000} & \textbf{1.000} & 0.997 & \textbf{1.000} & 0.997 \\
			glass-0-1-6\_vs\_2 & \textbf{0.771} & 0.692 & 0.733 & 0.725 & 0.707 & 0.524 & 0.685 & 0.646 \\
			glass2 & 0.737 & 0.717 & \textbf{0.839} & 0.805 & 0.801 & 0.779 & 0.701 & 0.666 \\
			yeast-1\_vs\_7 & \textbf{0.710} & 0.663 & 0.595 & 0.654 & 0.608 & 0.614 & 0.681 & 0.691 \\
			glass4 & 0.795 & 0.871 & 0.846 & 0.850 & 0.859 & \textbf{0.892} & 0.811 & 0.819 \\
			ecoli4 & \textbf{0.909} & 0.841 & 0.893 & 0.883 & 0.883 & 0.863 & 0.893 & 0.893 \\
			page-blocks-1-3\_vs\_4 & 0.982 & 0.944 & 0.964 & 0.972 & 0.964 & 0.982 & \textbf{0.990} & \textbf{0.990} \\
			abalone9-18 & 0.777 & 0.763 & 0.760 & 0.767 & 0.773 & 0.752 & \textbf{0.817} & 0.792 \\
			yeast-1-4-5-8\_vs\_7 & 0.593 & \textbf{0.637} & 0.584 & 0.618 & 0.628 & 0.487 & 0.489 & 0.489 \\
			glass5 & 0.912 & 0.843 & 0.792 & \textbf{0.919} & 0.780 & 0.792 & 0.633 & 0.633 \\
			yeast-2\_vs\_8 & \textbf{0.884} & 0.758 & 0.746 & 0.772 & 0.750 & 0.823 & 0.876 & 0.876 \\
			car\_eval\_4 & \textbf{1.000} & 0.967 & 0.997 & 0.994 & 0.997 & 0.993 & \textbf{1.000} & \textbf{1.000} \\
			wine\_quality & \textbf{0.753} & 0.681 & 0.660 & 0.690 & 0.674 & 0.669 & 0.675 & 0.674 \\
			yeast\_me2 & 0.701 & 0.638 & 0.668 & 0.655 & 0.656 & 0.690 & \textbf{0.707} & 0.702 \\
			yeast4 & \textbf{0.793} & 0.698 & 0.682 & 0.690 & 0.671 & 0.738 & 0.752 & 0.752 \\
			yeast-1-2-8-9\_vs\_7 & \textbf{0.775} & 0.642 & 0.612 & 0.633 & 0.607 & 0.698 & 0.750 & 0.750 \\
			yeast5 & 0.667 & 0.854 & 0.871 & 0.877 & \textbf{0.881} & 0.786 & 0.839 & 0.844 \\
			yeast6 & \textbf{0.745} & 0.734 & 0.730 & 0.724 & 0.705 & 0.696 & 0.708 & 0.738 \\
			abalone19 & 0.498 & 0.500 & \textbf{0.526} & 0.518 & 0.524 & 0.497 & 0.498 & 0.498 \\
			\bottomrule
		\end{tabular}%
	}
	\label{tab:F1AllDatasets}%
\end{table}%

\begin{table}[!htbp]
	\centering
	\caption{AUC result over different datasets.}
	\resizebox{0.97\columnwidth}{!}{%
		
		\begin{tabular}{lrrrrrrrr}
			\toprule
			& \multicolumn{1}{c}{SIMPOR} & \multicolumn{1}{c}{GDO} & \multicolumn{1}{c}{SMOTE} & \multicolumn{1}{c}{BL-SMOTE} & \multicolumn{1}{c}{ADASYN} & \multicolumn{1}{c}{DeepSM} & \multicolumn{1}{c}{SVMCS} & \multicolumn{1}{c}{EE} \\
			\midrule
			glass1 & 0.798 & \textbf{0.818} & 0.788 & 0.782 & 0.807 & 0.804 & 0.800 & 0.795 \\
			wisconsin & \textbf{0.995} & 0.992 & 0.992 & 0.992 & 0.992 & 0.991 & \textbf{0.995} & 0.994 \\
			pima  & \textbf{0.858} & 0.790 & 0.800 & 0.804 & 0.782 & 0.810 & 0.826 & 0.818 \\
			glass0 & \textbf{0.901} & 0.879 & 0.885 & 0.859 & 0.873 & 0.891 & 0.896 & 0.882 \\
			yeast1 & \textbf{0.811} & 0.758 & 0.753 & 0.753 & 0.746 & 0.776 & 0.782 & 0.774 \\
			haberman & 0.675 & 0.662 & 0.660 & 0.673 & 0.667 & 0.686 & 0.686 & \textbf{0.689} \\
			vehicle1 & \textbf{0.936} & 0.917 & 0.922 & 0.920 & 0.929 & 0.924 & 0.920 & 0.928 \\
			vehicle2 & \textbf{0.999} & 0.998 & \textbf{0.999} & \textbf{0.999} & \textbf{0.999} & 0.991 & \textbf{0.999} & \textbf{0.999} \\
			vehicle3 & \textbf{0.918} & 0.871 & 0.895 & 0.897 & 0.900 & 0.901 & 0.903 & 0.904 \\
			creditcard & \textbf{0.974} & 0.969 & 0.966 & 0.962 & 0.962 & 0.961 & 0.968 & 0.945 \\
			glass-0-1-2-3\_vs\_4-5-6 & 0.968 & 0.987 & \textbf{0.989} & 0.976 & 0.988 & 0.987 & 0.985 & 0.985 \\
			vehicle0 & 0.975 & 0.991 & 0.995 & 0.995 & \textbf{0.996} & 0.992 & 0.995 & \textbf{0.996} \\
			ecoli1 & 0.949 & 0.948 & \textbf{0.952} & 0.942 & 0.943 & 0.951 & 0.951 & 0.950 \\
			new-thyroid1 & \textbf{0.999} & \textbf{0.999} & 0.997 & 0.997 & 0.997 & 0.982 & 0.997 & 0.997 \\
			new-thyroid2 & \textbf{0.999} & \textbf{0.999} & 0.998 & 0.998 & 0.997 & 0.977 & 0.998 & \textbf{0.999} \\
			ecoli2 & 0.950 & 0.953 & 0.957 & 0.946 & 0.958 & 0.957 & 0.959 & \textbf{0.960} \\
			glass6 & \textbf{0.963} & 0.960 & 0.920 & 0.833 & 0.841 & 0.894 & 0.939 & 0.877 \\
			yeast3 & \textbf{0.968} & 0.943 & 0.935 & 0.927 & 0.937 & 0.946 & 0.966 & 0.967 \\
			ecoli3 & 0.883 & 0.879 & 0.878 & 0.880 & 0.883 & 0.891 & \textbf{0.897} & 0.885 \\
			page-blocks0 & \textbf{0.990} & 0.986 & 0.969 & 0.982 & 0.984 & 0.981 & 0.986 & 0.986 \\
			yeast-2\_vs\_4 & 0.974 & \textbf{0.976} & 0.961 & 0.959 & 0.960 & 0.907 & 0.972 & 0.949 \\
			yeast-0-5-6-7-9\_vs\_4 & 0.915 & \textbf{0.923} & 0.881 & 0.904 & 0.866 & 0.876 & 0.918 & 0.914 \\
			vowel0 & \textbf{1.000} & \textbf{1.000} & \textbf{1.000} & \textbf{1.000} & \textbf{1.000} & \textbf{1.000} & \textbf{1.000} & \textbf{1.000} \\
			glass-0-1-6\_vs\_2 & \textbf{0.942} & 0.897 & 0.905 & 0.892 & 0.910 & 0.886 & 0.907 & 0.941 \\
			glass2 & 0.929 & 0.917 & 0.923 & 0.923 & \textbf{0.952} & 0.919 & 0.940 & 0.932 \\
			yeast-1\_vs\_7 & \textbf{0.848} & 0.777 & 0.677 & 0.761 & 0.685 & 0.702 & 0.791 & 0.795 \\
			glass4 & 0.955 & 0.976 & 0.954 & \textbf{0.987} & 0.949 & 0.979 & 0.972 & 0.975 \\
			ecoli4 & \textbf{0.997} & 0.978 & 0.984 & 0.984 & 0.989 & 0.953 & 0.990 & 0.988 \\
			page-blocks-1-3\_vs\_4 & \textbf{1.000} & 0.997 & 0.999 & 0.999 & 0.999 & 0.999 & 0.999 & 0.999 \\
			abalone9-18 & 0.934 & 0.920 & 0.933 & 0.929 & 0.919 & 0.898 & 0.930 & \textbf{0.940} \\
			yeast-1-4-5-8\_vs\_7 & \textbf{0.823} & 0.746 & 0.721 & 0.734 & 0.721 & 0.721 & 0.769 & 0.754 \\
			glass5 & 0.987 & \textbf{0.990} & 0.985 & 0.983 & 0.987 & 0.987 & \textbf{0.990} & 0.988 \\
			yeast-2\_vs\_8 & 0.855 & 0.845 & 0.835 & \textbf{0.865} & 0.853 & 0.802 & 0.809 & 0.805 \\
			car\_eval\_4 & \textbf{1.000} & \textbf{1.000} & \textbf{1.000} & \textbf{1.000} & \textbf{1.000} & \textbf{1.000} & \textbf{1.000} & \textbf{1.000} \\
			wine\_quality & \textbf{0.852} & 0.783 & 0.727 & 0.756 & 0.740 & 0.793 & 0.781 & 0.805 \\
			yeast\_me2 & \textbf{0.896} & 0.887 & 0.793 & 0.817 & 0.787 & 0.832 & 0.871 & 0.874 \\
			yeast4 & \textbf{0.935} & 0.889 & 0.796 & 0.810 & 0.792 & 0.817 & 0.832 & 0.848 \\
			yeast-1-2-8-9\_vs\_7 & \textbf{0.761} & 0.699 & 0.702 & 0.687 & 0.685 & 0.695 & 0.745 & 0.756 \\
			yeast5 & 0.835 & 0.991 & 0.985 & 0.984 & 0.985 & 0.983 & 0.992 & \textbf{0.993} \\
			yeast6 & 0.960 & 0.936 & 0.905 & 0.946 & 0.906 & 0.933 & 0.959 & \textbf{0.963} \\
			abalone19 & \textbf{0.782} & 0.557 & 0.616 & 0.676 & 0.575 & 0.721 & 0.763 & 0.771 \\
			\bottomrule
		\end{tabular}%
		
	}
	\label{tab:AUCAllDatasets}%
\end{table}%

\begin{table}[!htbp]
	\centering
	\caption{Recall results over 41 datasets.}
	\resizebox{0.97\columnwidth}{!}{%
		\begin{tabular}{lrrrrrrrr}
			\toprule
			& \multicolumn{1}{c}{SIMPOR} & \multicolumn{1}{c}{GDO} & \multicolumn{1}{c}{SMOTE} & \multicolumn{1}{c}{BL-SMOTE} & \multicolumn{1}{c}{ADASYN} & \multicolumn{1}{c}{DeepSM} & \multicolumn{1}{c}{SVMCS} & \multicolumn{1}{c}{EE} \\
			\midrule
			glass1 & 0.725 & \textbf{0.739} & 0.705 & 0.726 & 0.727 & 0.702 & 0.713 & 0.699 \\
			wisconsin & \textbf{0.965} & 0.963 & 0.955 & 0.961 & 0.959 & 0.962 & 0.960 & 0.958 \\
			pima  & \textbf{0.778} & 0.705 & 0.714 & 0.727 & 0.701 & 0.715 & 0.733 & 0.726 \\
			glass0 & 0.844 & 0.817 & 0.814 & 0.810 & 0.818 & 0.817 & \textbf{0.847} & 0.815 \\
			yeast1 & \textbf{0.704} & 0.679 & 0.680 & 0.691 & 0.677 & 0.660 & 0.669 & 0.668 \\
			haberman & 0.593 & \textbf{0.611} & 0.601 & 0.599 & 0.590 & 0.583 & 0.570 & 0.563 \\
			vehicle1 & 0.814 & \textbf{0.827} & 0.803 & 0.789 & 0.816 & 0.774 & 0.777 & 0.800 \\
			vehicle2 & \textbf{0.988} & 0.980 & 0.983 & 0.983 & 0.985 & 0.965 & 0.982 & 0.986 \\
			vehicle3 & \textbf{0.817} & 0.772 & 0.791 & 0.787 & 0.806 & 0.772 & 0.770 & 0.773 \\
			creditcard & \textbf{0.948} & 0.937 & 0.937 & 0.936 & 0.935 & 0.936 & 0.911 & 0.925 \\
			glass-0-1-2-3\_vs\_4-5-6 & 0.857 & 0.932 & 0.915 & 0.910 & 0.920 & \textbf{0.935} & 0.909 & 0.901 \\
			vehicle0 & 0.947 & \textbf{0.979} & \textbf{0.979} & 0.971 & 0.973 & 0.946 & 0.969 & 0.969 \\
			ecoli1 & 0.854 & 0.852 & 0.866 & 0.847 & 0.849 & \textbf{0.880} & 0.837 & 0.839 \\
			new-thyroid1 & 0.965 & \textbf{0.995} & 0.942 & 0.953 & 0.953 & 0.901 & 0.942 & 0.942 \\
			new-thyroid2 & 0.951 & \textbf{0.995} & 0.913 & 0.913 & 0.913 & 0.859 & 0.900 & 0.900 \\
			ecoli2 & \textbf{0.916} & 0.913 & 0.911 & 0.878 & 0.902 & 0.910 & 0.903 & 0.905 \\
			glass6 & \textbf{0.931} & 0.880 & 0.818 & 0.800 & 0.820 & 0.816 & 0.800 & 0.800 \\
			yeast3 & 0.856 & 0.863 & 0.862 & 0.860 & 0.878 & 0.834 & 0.869 & \textbf{0.883} \\
			ecoli3 & 0.823 & \textbf{0.870} & 0.850 & 0.850 & 0.861 & 0.855 & 0.816 & 0.814 \\
			page-blocks0 & 0.923 & \textbf{0.955} & 0.924 & 0.932 & 0.952 & 0.892 & 0.917 & 0.913 \\
			yeast-2\_vs\_4 & 0.895 & \textbf{0.908} & 0.881 & 0.827 & 0.868 & 0.810 & 0.875 & 0.741 \\
			yeast-0-5-6-7-9\_vs\_4 & 0.814 & \textbf{0.832} & 0.770 & 0.815 & 0.802 & 0.771 & 0.772 & 0.752 \\
			vowel0 & \textbf{1.000} & \textbf{1.000} & \textbf{1.000} & \textbf{1.000} & \textbf{1.000} & 0.995 & \textbf{1.000} & 0.995 \\
			glass-0-1-6\_vs\_2 & \textbf{0.741} & 0.680 & 0.735 & 0.735 & 0.730 & 0.520 & 0.716 & 0.682 \\
			glass2 & 0.747 & 0.800 & \textbf{0.939} & 0.869 & 0.907 & 0.812 & 0.677 & 0.689 \\
			yeast-1\_vs\_7 & 0.655 & \textbf{0.722} & 0.611 & 0.678 & 0.634 & 0.562 & 0.633 & 0.635 \\
			glass4 & 0.841 & \textbf{0.906} & 0.795 & 0.795 & 0.872 & 0.821 & 0.759 & 0.761 \\
			ecoli4 & \textbf{0.934} & 0.901 & 0.911 & 0.909 & 0.909 & 0.928 & 0.911 & 0.911 \\
			page-blocks-1-3\_vs\_4 & 0.998 & 0.992 & 0.996 & 0.997 & 0.996 & 0.998 & \textbf{0.999} & \textbf{0.999} \\
			abalone9-18 & 0.751 & \textbf{0.830} & 0.812 & 0.812 & 0.823 & 0.748 & 0.751 & 0.733 \\
			yeast-1-4-5-8\_vs\_7 & 0.549 & \textbf{0.723} & 0.618 & 0.665 & 0.697 & 0.495 & 0.499 & 0.499 \\
			glass5 & \textbf{0.912} & 0.868 & 0.768 & 0.893 & 0.766 & 0.768 & 0.602 & 0.602 \\
			yeast-2\_vs\_8 & \textbf{0.872} & 0.835 & 0.858 & 0.838 & 0.858 & 0.845 & 0.848 & 0.848 \\
			car\_eval\_4 & \textbf{1.000} & 0.997 & \textbf{1.000} & 0.999 & \textbf{1.000} & 0.986 & \textbf{1.000} & \textbf{1.000} \\
			wine\_quality & \textbf{0.750} & 0.706 & 0.639 & 0.670 & 0.667 & 0.620 & 0.634 & 0.623 \\
			yeast\_me2 & 0.671 & \textbf{0.693} & 0.670 & 0.658 & 0.673 & 0.647 & 0.610 & 0.609 \\
			yeast4 & \textbf{0.790} & 0.776 & 0.732 & 0.725 & 0.722 & 0.741 & 0.690 & 0.690 \\
			yeast-1-2-8-9\_vs\_7 & 0.685 & \textbf{0.694} & 0.641 & 0.649 & 0.627 & 0.602 & 0.622 & 0.604 \\
			yeast5 & 0.716 & \textbf{0.979} & 0.955 & 0.964 & 0.956 & 0.830 & 0.883 & 0.883 \\
			yeast6 & 0.714 & \textbf{0.827} & 0.813 & 0.723 & 0.776 & 0.691 & 0.681 & 0.691 \\
			abalone19 & 0.499 & 0.503 & 0.537 & 0.518 & \textbf{0.539} & 0.497 & 0.500 & 0.500 \\
			\bottomrule
		\end{tabular}%
		
	}
	\label{tab:Recall}%
\end{table}%

\begin{table}[!hhbp]
	\centering
	\caption{Precision results over 41 datasets.}
	\resizebox{0.97\columnwidth}{!}{%
		
		\begin{tabular}{lrrrrrrrr}
			\toprule
			& \multicolumn{1}{c}{SIMPOR} & \multicolumn{1}{c}{GDO} & \multicolumn{1}{c}{SMOTE} & \multicolumn{1}{c}{BL-SMOTE} & \multicolumn{1}{c}{ADASYN} & \multicolumn{1}{c}{DeepSM} & \multicolumn{1}{c}{SVMCS} & \multicolumn{1}{c}{EE} \\
			\midrule
			glass1 & 0.733 & \textbf{0.744} & 0.710 & 0.732 & 0.730 & 0.710 & 0.726 & 0.711 \\
			wisconsin & \textbf{0.959} & 0.955 & 0.952 & 0.956 & 0.954 & 0.957 & 0.956 & 0.955 \\
			pima  & \textbf{0.776} & 0.694 & 0.714 & 0.715 & 0.700 & 0.727 & 0.752 & 0.737 \\
			glass0 & \textbf{0.836} & 0.782 & 0.796 & 0.782 & 0.795 & 0.809 & 0.825 & 0.808 \\
			yeast1 & \textbf{0.727} & 0.674 & 0.670 & 0.680 & 0.668 & 0.687 & 0.702 & 0.689 \\
			haberman & \textbf{0.610} & 0.588 & 0.578 & 0.574 & 0.570 & 0.592 & 0.603 & 0.608 \\
			vehicle1 & \textbf{0.835} & 0.803 & 0.811 & 0.804 & 0.819 & 0.797 & 0.793 & 0.817 \\
			vehicle2 & \textbf{0.986} & 0.954 & 0.970 & 0.973 & 0.977 & 0.944 & 0.971 & 0.977 \\
			vehicle3 & \textbf{0.826} & 0.760 & 0.779 & 0.797 & 0.807 & 0.788 & 0.794 & 0.798 \\
			creditcard & \textbf{0.961} & 0.933 & 0.955 & 0.953 & 0.951 & 0.958 & 0.904 & 0.954 \\
			glass-0-1-2-3\_vs\_4-5-6 & 0.845 & 0.915 & 0.922 & 0.915 & 0.911 & \textbf{0.925} & 0.907 & 0.911 \\
			vehicle0 & 0.921 & 0.935 & 0.960 & 0.960 & 0.956 & 0.959 & \textbf{0.971} & 0.969 \\
			ecoli1 & 0.810 & 0.795 & 0.811 & 0.792 & 0.784 & \textbf{0.827} & 0.813 & 0.816 \\
			new-thyroid1 & \textbf{0.977} & 0.964 & 0.950 & 0.953 & 0.953 & 0.903 & 0.950 & 0.950 \\
			new-thyroid2 & \textbf{0.974} & 0.971 & 0.966 & 0.966 & 0.966 & 0.886 & 0.963 & 0.963 \\
			ecoli2 & \textbf{0.928} & 0.849 & 0.901 & 0.852 & 0.867 & 0.867 & 0.915 & 0.924 \\
			glass6 & 0.977 & 0.922 & 0.941 & \textbf{0.980} & 0.926 & 0.920 & \textbf{0.980} & \textbf{0.980} \\
			yeast3 & 0.868 & 0.777 & 0.825 & 0.815 & 0.786 & 0.829 & 0.867 & \textbf{0.876} \\
			ecoli3 & 0.791 & 0.726 & 0.739 & 0.743 & 0.734 & 0.806 & \textbf{0.842} & 0.838 \\
			page-blocks0 & 0.929 & 0.858 & 0.897 & 0.870 & 0.854 & \textbf{0.936} & 0.922 & 0.913 \\
			yeast-2\_vs\_4 & 0.874 & 0.846 & \textbf{0.907} & 0.864 & 0.865 & 0.830 & 0.750 & 0.818 \\
			yeast-0-5-6-7-9\_vs\_4 & 0.835 & 0.688 & 0.741 & 0.755 & 0.721 & 0.748 & 0.861 & \textbf{0.868} \\
			vowel0 & \textbf{1.000} & \textbf{1.000} & \textbf{1.000} & \textbf{1.000} & \textbf{1.000} & 0.999 & \textbf{1.000} & 0.999 \\
			glass-0-1-6\_vs\_2 & \textbf{0.822} & 0.710 & 0.760 & 0.728 & 0.720 & 0.531 & 0.664 & 0.619 \\
			glass2 & 0.731 & 0.653 & \textbf{0.765} & 0.764 & 0.723 & 0.758 & 0.739 & 0.649 \\
			yeast-1\_vs\_7 & \textbf{0.779} & 0.613 & 0.580 & 0.632 & 0.584 & 0.693 & 0.740 & 0.759 \\
			glass4 & 0.756 & 0.845 & 0.910 & 0.927 & 0.853 & \textbf{0.985} & 0.887 & 0.899 \\
			ecoli4 & \textbf{0.896} & 0.792 & 0.879 & 0.862 & 0.862 & 0.808 & 0.879 & 0.879 \\
			page-blocks-1-3\_vs\_4 & 0.967 & 0.903 & 0.935 & 0.949 & 0.935 & 0.969 & \textbf{0.982} & \textbf{0.982} \\
			abalone9-18 & 0.806 & 0.706 & 0.717 & 0.731 & 0.731 & 0.762 & \textbf{0.903} & 0.866 \\
			yeast-1-4-5-8\_vs\_7 & \textbf{0.668} & 0.574 & 0.557 & 0.582 & 0.579 & 0.479 & 0.479 & 0.479 \\
			glass5 & 0.912 & 0.823 & 0.824 & \textbf{0.954} & 0.803 & 0.824 & 0.673 & 0.673 \\
			yeast-2\_vs\_8 & 0.913 & 0.728 & 0.670 & 0.752 & 0.682 & 0.825 & \textbf{0.921} & \textbf{0.921} \\
			car\_eval\_4 & \textbf{1.000} & 0.939 & 0.994 & 0.988 & 0.994 & 0.999 & \textbf{1.000} & \textbf{1.000} \\
			wine\_quality & \textbf{0.758} & 0.659 & 0.682 & 0.713 & 0.682 & 0.727 & 0.723 & 0.736 \\
			yeast\_me2 & 0.737 & 0.593 & 0.666 & 0.652 & 0.641 & 0.738 & \textbf{0.852} & 0.843 \\
			yeast4 & 0.800 & 0.635 & 0.640 & 0.660 & 0.629 & 0.738 & 0.829 & \textbf{0.831} \\
			yeast-1-2-8-9\_vs\_7 & 0.909 & 0.598 & 0.587 & 0.620 & 0.589 & 0.842 & 0.957 & \textbf{0.988} \\
			yeast5 & 0.629 & 0.759 & 0.804 & 0.807 & \textbf{0.821} & 0.753 & 0.807 & 0.816 \\
			yeast6 & 0.785 & 0.662 & 0.671 & 0.729 & 0.657 & 0.714 & 0.740 & \textbf{0.801} \\
			abalone19 & 0.497 & 0.498 & 0.516 & \textbf{0.517} & 0.511 & 0.497 & 0.497 & 0.497 \\
			\bottomrule
		\end{tabular}%
		
	}
	\label{tab:Precision}%
\end{table}%

\newpage

\section{Empirical Study}

\begin{figure*}[ht]
	\centering
	\includegraphics[width=0.98\linewidth, trim=5 10 10 10, clip ]{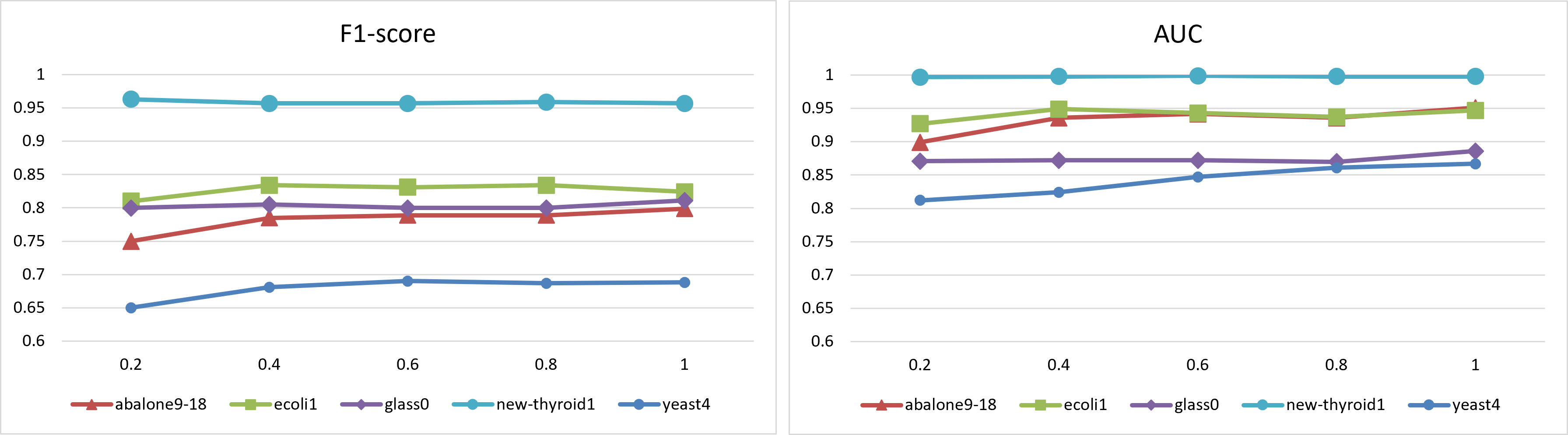}
	\caption{F1-score and AUC results with varying Gaussian standard deviation (ranging from 0.2R to R).}
	\label{fig:r_result}
\end{figure*}

\begin{figure*}[h]
	\centering
	\includegraphics[width=0.98\linewidth, trim=5 10 10 10, clip ]{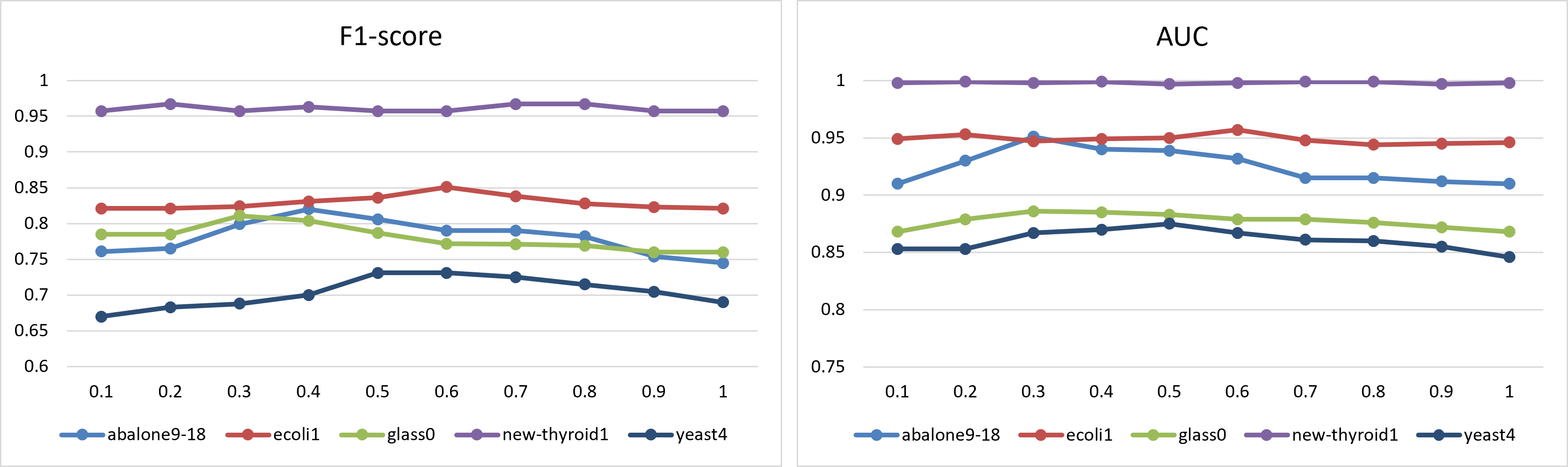}
	\caption{F1-score and AUC results with varying informative portion IP.}
	\label{fig:ip_result}
\end{figure*}

\subsection{Empirical study on the impact of radius factor r.}
\label{sec:simpor_r_distribution_impact}
In this section, we study how the classification performance is impacted by different generation radius factor $r$ in Equation \ref{equ:r_dist}. The classification performance is measured under different distribution settings of the radius r as it controls how far synthetic data are generated from its original minority sample. We use different parameters for the Gaussian distribution $\mathcal{N}(\mu ,\,{(\alpha R)}^{2})$. Particularly, we fix the mean value to zero and change $\alpha$ from 0.2 to 1 with steps of 0.2 so that the Gaussian standard deviation $\alpha R$ will range from 0.2R to R. To save space, we arbitrarily select 5 datasets to conduct this experiment. The classification results are shown in Figure \ref{fig:r_result}.

The figure obtained from the experiment indicates that the r factor, with a radius distribution standard deviation ranging from 0.6R to R, has minimal impact on the classification performance. While there are slight variations within the $\alpha$ range of 0.6 to 1, the performance improves between 0.2 to 0.6 (such as for ecoli1, abalone9-18, and yeast4). This is because the performance mainly depends on the classifier's decision boundary, and the synthetic data are placed far away from the decision boundary towards the minority class area; thus,  the radius does not have much effect on the accuracy results. However, in the case of multi-classed data, the performance might be affected by a significant value of R. 

\subsection{Empirical study on the impact of informative portion (IP).}
This section studies the empirical impact of the informative portion (IP) in Section \ref{sec:EAL}. This portion works as a threshold to adjust how many samples are taken into consideration of informative samples. To save space, we study five datasets used in Section \ref{sec:simpor_r_distribution_impact}. Different values of IP ranging from 0.1 to 1 are applied, and the classification performance results are shown in Figure \ref{fig:ip_result}.

As we can see from the figure, while datasets with outstanding performance (new-thyroid1, ecoli1) have little impact, there are fluctuations in other datasets' F1-score and AUC score  (abalonce9-18, glass0, yeast4). This is because, for the easy-separated dataset such as new-thyroid1 and ecoli1, the IP change does not affect the classification performance as the data classes are easily separated. While in more challenging datasets, IP changes might affect the balance at the informative region; thus, this leads to performance variations. The resulting figure also suggests tuning IP for each dataset between a range of (0.2, 0.6) could achieve higher performance.

\newpage

\section{Data visualization}  

To explore more on how the techniques perform, we visualize the generated data by projecting them onto lower dimension space (i.e., one and two dimensions) using the Principle Component Analysis technique (PCA) \cite{pca}. Data's 2-Dimension (2D) plots and 1-Dimension histograms are presented with a hard-to-differentiate ratio (HDR) for each technique. 1D histograms are computed by dividing one-dimensional-reduced data into 20 bins (intervals) and counting the number of samples within the interval of each bin. A hard-to-differentiate ratio is defined as the ratio of the number of samples in the intersection between 2 classes to the total of minority samples ($HDR = \frac{No.\: Intersection \: samples}{No. \: Minority \: samples}100\% $) where the number of intersection samples is estimated by counting samples in the overlapped bins between the two classes in the 1D histograms. This ratio is expected to be as small as 0\% if the two classes are well separated; in contrast, 100\% indicates that the two classes cannot be distinguished in the projected 1D space. Besides HDR, we show the absolute numbers of Minority, Majority, and Intersection samples for each technique in the bottom tables. From the plots, we observe how the data are distributed in 2D space and quantify samples that are hard to be differentiated in the 1D space histograms.

\begin{figure*}[h!]
	\centering
	\includegraphics[width=0.8\linewidth,trim=10 90 10 10, clip ]{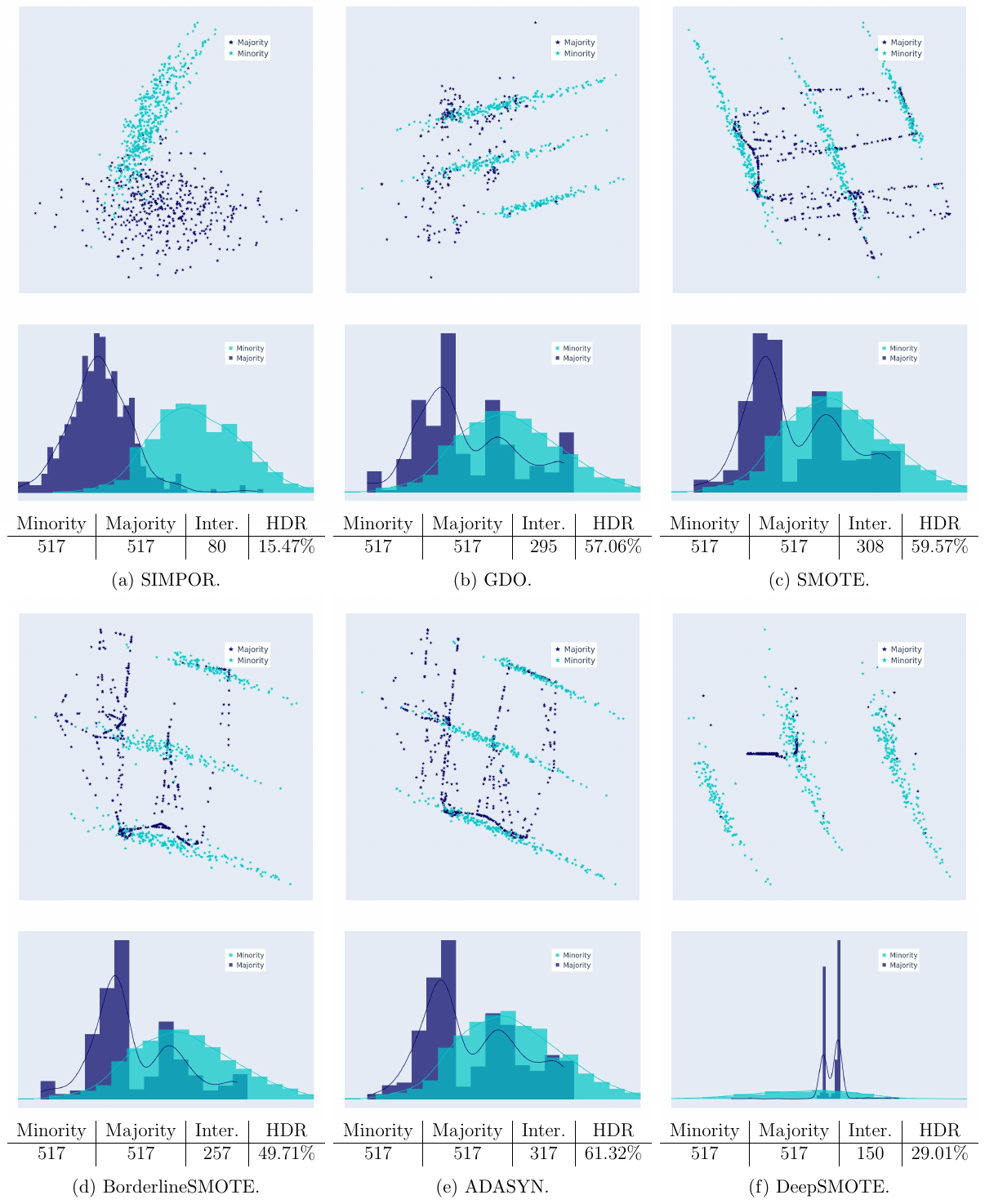}
	\caption{Abalone9-18: Generated training data projected onto 2-dimension space and their histograms in 1-Dimension space using Principle Component Analysis dimension reduction technique. The bottom tables illustrate the number of samples in two classes, 1-Dimension histogram intersection between 2 classes, and the hard-to-differentiate ratio between the number of intersection samples to the number of minority samples ($HDR = \frac{Inter.}{Minority}100\%$).}
	\label{fig:visualization1d2d}
\end{figure*}

To save space, we only show the plot of one dataset (i.e., Abalone9-18 dataset) in Figure \ref{fig:visualization1d2d}. Many other datasets are observed to have similar patterns. We observe that the proposed technique does not poorly generate synthetic samples as many as other techniques do. HDR results show that \Methodname{} achieves the least number of hard-to-differentiate ratio at 15.47\%. As shown in the 2D visualization sub-figures, other techniques poorly-place synthetic data crossed the other class. This causes by outliers or noises near the border between the two classes that other techniques do not pay attention to and mistakenly create more noise. In contrast, \Methodname{} safely produces synthetic data towards the minority class by maximizing the posterior ratio ;thus it can reduce the number of poorly-placed samples.

\newpage

\section{Datasets Description}
\label{apx.dataDecription}

\begin{table}[!htbp]
	\centering
	\caption{Dataset Description.}
	
	\begin{tabular}{lccc}
		\hline
		dataset & \#samples & \#features & IR \bigstrut\\
		\hline
		glass1 & 214   & 9     & 1.8 (138:76)\bigstrut[t]\\
		wisconsin & 683   & 9     & 1.9 (444:239) \\
		pima  & 768   & 8     & 1.9 (500:268)\\
		glass0 & 214   & 9     & 2.1 (144:70) \\
		yeast1 & 1484  & 8     & 2.5 (1055:429)\\
		haberman & 306   & 3     & 2.8 (225:81) \\
		vehicle1 & 846   & 18    & 2.9 (629:217)\\
		vehicle2 & 846   & 18    & 2.9 (628:218)\\
		vehicle3 & 846   & 18    & 3.0 (634:212)\\
		creditcard & 1968  & 30    & 3.0 (1476:492)\\
		glass-0-1-2-3\_vs\_4-5-6 & 214   & 9     & 3.2 (163:51)\\
		vehicle0 & 846   & 18    & 3.3 (647:199)\\
		ecoli1 & 336   & 7     & 3.4 (259:77)\\
		new-thyroid1 & 215   & 5     & 5.1 (180:35) \\
		new-thyroid2 & 215   & 5     & 5.1 (180:35)\\
		ecoli2 & 336   & 7     & 5.5 (284:52)\\
		glass6 & 214   & 9     & 6.4 (185:29)\\
		yeast3 & 1484  & 8     & 8.1 (1321:63)\\
		ecoli3 & 336   & 7     & 8.6 (301:35)\\
		page-blocks0 & 5472  & 10    & 8.8 (4913:559)\\
		yeast-2\_vs\_4 & 514   & 8     & 9.0 (463:51)\\
		yeast-0-5-6-7-9\_vs\_4 & 528   & 8     & 9.4 (477:51)\\
		vowel0 & 988   & 13    & 10.0 (898:90)\\
		glass-0-1-6\_vs\_2 & 192   & 9     & 10.3 (175:17)\\
		glass2 & 214   & 9     & 11.6 (197:17)\\
		yeast-1\_vs\_7 & 459   & 7     & 14.3 (429:30)\\
		glass4 & 214   & 9     & 15.5 (201:13)\\
		ecoli4 & 336   & 7     & 15.8 (316:20)\\
		page-blocks-1-3\_vs\_4 & 472   & 10    & 15.9 (444:28)\\
		abalone9-18 & 731   & 8     & 16.4 (689:42)\\
		yeast-1-4-5-8\_vs\_7 & 693   & 8     & 22.1 (663:30)\\
		glass5 & 214   & 9     & 22.8 (205:9)\\
		yeast-2\_vs\_8 & 482   & 8     & 23.1 (462:20)\\
		car\_eval\_4 & 1728  & 21    & 25.6 (1663:65)\\
		wine\_quality & 4898  & 11    & 25.8 (4715:183)\\
		yeast\_me2 & 1484  & 8     & 28.0 (1433:51)\\
		yeast4 & 1484  & 8     & 28.1 (1433:51)\\
		yeast-1-2-8-9\_vs\_7 & 947   & 8     & 30.6 (917:30)\\
		yeast5 & 1484  & 8     & 32.7 (1440:44)\\
		yeast6 & 1484  & 8     & 41.4 (1449:35)\\
		abalone19 & 4174  & 8     & 129.4 (689:42)\\
	\end{tabular}%
	\label{tab:dataDecription}%
\end{table}%

\end{document}